\documentclass[sigconf]{acmart}
\AtBeginDocument{%
  }

\usepackage[inkscapelatex=false]{svg}
\usepackage{enumitem}
\usepackage{soul}
\usepackage{graphicx}
\usepackage{caption}
\usepackage{subcaption}
\usepackage{amsmath} 
\usepackage{tabularx}
\usepackage{multirow}

\usepackage{multicol}
\usepackage{svg}
\usepackage{graphicx}
\usepackage{booktabs}
\usepackage{geometry}
\newtheorem{definition}{\textbf{Definition}}

\copyrightyear{2025}
\acmYear{2025}
\setcopyright{cc}
\setcctype{by}
\acmConference[SIGSPATIAL '25]{The 33rd ACM International Conference on Advances in Geographic Information Systems}{November 3--6, 2025}{Minneapolis, MN, USA}
\acmBooktitle{The 33rd ACM International Conference on Advances in Geographic Information Systems (SIGSPATIAL '25), November 3--6, 2025, Minneapolis, MN, USA}
\acmDOI{10.1145/3748636.3764173}
\acmISBN{979-8-4007-2086-4/2025/11}

\begin{document}

\title[MuST$^2$-Learn]{MuST$^2$-Learn: Multi-view Spatial-Temporal-Type Learning for Heterogeneous Municipal Service Time Estimation}

\author{Nadia Asif}
\affiliation{%
  \institution{University of Tennessee at}
  \city{Chattanooga}
  \state{TN}
  \country{USA}}
\email{hdh898@mocs.utc.edu}

\author{Zhiqing Hong}
\affiliation{%
  \institution{Rutgers University}
  \city{Piscataway}
  \state{NJ}
  \country{USA}}
\email{zhiqing.hong@rutgers.edu}

\author{Shaogang Ren}
\affiliation{%
  \institution{University of Tennessee at}
  \city{Chattanooga}
  \state{TN}
  \country{USA}}
\email{shaogang-ren@utc.edu}

\author{Xiaonan Zhang}
\affiliation{%
  \institution{Florida State University}
  \city{Tallahassee}
  \state{FL}
  \country{USA}}
\email{xzhang14@fsu.edu}

\author{Xiaojun Shang}
\affiliation{%
  \institution{University of Texas at Arlington}
  \city{Arlington}
  \state{TX}
  \country{USA}}
\email{xiaojun.shang@uta.edu}

\author{Yukun Yuan}
\affiliation{%
  \institution{University of Tennessee at}
  \city{Chattanooga}
  \state{TN}
  \country{USA}}
\email{yukun-yuan@utc.edu
}

\renewcommand{\shortauthors}{N. Asif et al.}

\begin{abstract}
Non-emergency municipal services, e.g., city 311 systems, have been widely implemented across cities in Canada and the United States to enhance residents’ quality of life. These systems enable residents to report issues, e.g., noise complaints, missed garbage collection, and potholes, via phone calls, mobile applications, or webpages.
However, residents are often given limited information about when their service requests will be addressed, which can reduce transparency, lower resident satisfaction, and increase the number of follow-up inquiries.
Predicting the service time for municipal service requests is challenging due to several complex factors:
(i) dynamic spatial-temporal correlations,
(ii) underlying interactions among heterogeneous service request types, and
(iii) high variation in service duration even within the same request category.
In this work, we propose MuST$^2$-Learn: a \textbf{Mu}lti-view \textbf{S}patial-\textbf{T}emporal-\textbf{T}ype \textbf{Learn}ing framework designed to address the aforementioned challenges by jointly modeling spatial, temporal, and service type dimensions. 
In detail, it incorporates an inter-type encoder to capture relationships among heterogeneous service request types and an intra-type variation encoder to model service time variation within homogeneous types.
In addition, a spatiotemporal encoder is integrated to capture spatial and temporal correlations in each request type.
The proposed framework is evaluated with extensive experiments using two real-world datasets.
The results show that MuST$^2$-Learn reduces mean absolute error by at least 32.5\%, which outperforms state-of-the-art methods. 

\end{abstract}



\maketitle




\section{Introduction}

Non-emergency municipal service systems, such as the 311 platforms, have been widely adopted in cities across Canada and the United States to improve residents’ quality of life~\cite{wang2017structure,nyc311,chi311}.
These systems allow residents to report local issues, such as noise complaints, missed garbage collection, brush collection, and potholes, through phone calls, mobile applications, or webpages~\cite{nam2014understanding,hashemi2022automatic,holmes2007building}.
Once submitted, each service request is forwarded to the appropriate city department and added to its task queue for subsequent processing~\cite{wang2022community,nam2014understanding,10.1145/3463677.3463717}.
An important function of these service systems is to provide residents with an estimated completion date for their requests, similar to delivery time estimation~\cite{zhao2023hst} in online shopping and mailing services.
Providing estimated service times offers several benefits to both residents and city governments. 
These include improving government transparency, enhancing resident satisfaction, reducing resident inquiries and staff workload on communication, and supporting more effective planning and allocation of municipal services.

\begin{table*}[]
\centering
\caption{City department (division) and its assigned types of requests}
\label{table:departmentassignedtypes}
\vspace{-0.1in}
\small
\begin{tabular}{l|l}
\hline 
\multicolumn{1}{c|}{Department (division)} & \multicolumn{1}{c}{Request type examples} \\ \hline 
Public works (solid waste) & Bagged yard waste, brush collection, bulk trash, missed recycle, garbage container repair, missed garbage \\ \hline
Public works (street) & Loose leaf collection, roadside mowing, potholes, street sweeping, alley maintenance, distressed pavement \\ \hline
Public works (forestry) & Tree fallen/branch, tree trimming, tree removal, \\ \hline
Public works - stormwater & Erosion and drainage, flooding, storm drainage problem, storm grates \\ \hline
\begin{tabular}[c]{@{}l@{}}Economic community \\ development\end{tabular} & Illegal dumping, litter, house violation, abandoned or inoperable vehicle, graffiti removal, zone violation \\ \hline
Transportation department & Traffic signs, roadway sight obstruction, street lane markings, traffic lights \\ \hline
\end{tabular}
\end{table*}

Several studies have investigated city service time and delivery time estimation~\cite{wang2022community,10.1145/3286978.3287010,raj2021swift,9291583,zhao2023hst,zhou2023inductive,yi2023deepsta}.
\cite{wang2022community} applies the Prophet model to estimate the service time of specific 311 service calls by incorporating trend, seasonality, and holiday effects.
\cite{10.1145/3286978.3287010} and \cite{raj2021swift} develop structured regression models based on sparse Gaussian Conditional Random Fields to predict future response times.
\cite{zhao2023hst} proposes a heterogeneous spatial-temporal graph transformer for full-link delivery time estimation in warehouse-distribution systems.
In order to estimate package delivery time, \cite{zhou2023inductive} designs an inductive graph transformer, which integrates raw feature information and structural graph data.
We argue that a key limitation of these works is their treatment of heterogeneous request types as homogeneous, which prevents them from capturing latent inter-type and intra-type correlations that could enhance estimation accuracy.

In this work, we aim to integrate the heterogeneity across request types into our learning method to predict the service time of 311 requests.
There are several technical challenges of developing such a method.
First, service times are not independent across spatial and temporal dimensions. The service time for requests submitted in a given region on a specific day can be strongly influenced by the service times in neighboring regions and previous days.
Second, there exist latent interactions among heterogeneous request types. Multiple types of requests are often handled by the same department with limited resources, leading to potential cross-type interference. As a result, the service time of one request type may be affected by the demand level of other types.
Third, even within a single request type, service times can vary significantly due to individual request characteristics such as submission time, location, and workload. This intra-type variation reflects the complexity of service time estimation.
More analysis results to explain and justify these challenges are provided in Section~\ref{sec:challenges}.

To address the aforementioned challenges, in this paper, we propose MuST$^2$-Learn, a \textbf{Mu}lti-view \textbf{S}patial-\textbf{T}emporal-\textbf{T}ype \textbf{Learn}ing framework, which integrates the latent correlation from the spatial, temporal, and type dimensions.
In detail, we first develop an intra-type spatiotemporal encoder using LSTM and CNN to capture latent spatial-temporal patterns for each request type.
Second, to model correlations among heterogeneous request types within these patterns, we design an attention-based inter-type encoder that automatically learns the contribution of related types.
Third, we introduce an intra-type variation encoder based on Gaussian Process Regression and estimated request workload derived by feeding description of each request into a Large Language Model. This component captures service time variation within each type and the unique characteristics of each request, enhancing prediction robustness.
Finally, a multilayer perceptron is adopted to predict service time based on the outputs of the intra-type variation and inter-type encoders.


We have engaged with the City of Chattanooga to design a service time predictor and explore its integration into their 311 system, which currently does not provide estimated completion dates for submitted requests. Our goal is to offer this information to residents to enhance transparency and satisfaction. We are in the early stages of deployment, i.e., focusing on feasibility evaluation. In the future, we plan to extend the method and collaborate with additional cities.

Our main contributions are summarized as follows.
\begin{itemize}[leftmargin=*]
\item To the best of our knowledge, this is the first work to perform service time forecasting for municipal service requests considering the heterogeneity across request types and the service time variation of each request type. 
\item We design MuST$^2$-Learn: a multi-view learning framework that integrates intra-type spatiotemporal and variation encoders with an inter-type heterogeneity encoder to capture spatial, temporal, and type-specific patterns.
\item Extensive evaluations on two real-world datasets show that MuST$^2$-Learn outperforms state-of-the-art methods in service time prediction, achieving at least a 32.5\% reduction in mean absolute error.

\end{itemize}


\section{Background \& Challenges}



\subsection{311 Non-emergency Service Systems}

\begin{figure}[h]
    \centering
    \includegraphics[width=3.0in]{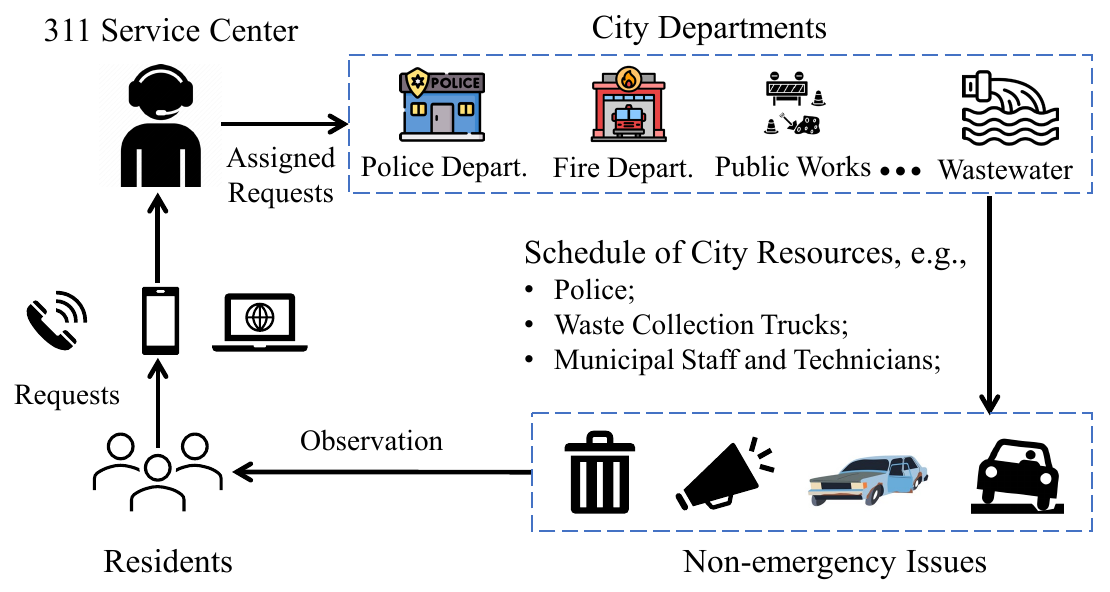}
    \vspace{-0.1in}
    \caption{Demonstration of City 311 Service Systems}
     \vspace{-0.1in}
    \label{fig:311framework}
\end{figure}

In many North American cities, the 311 system allows residents to report non-emergency issues and request municipal services, such as potholes, noise complaints, abandoned vehicles, or damaged traffic signs. As illustrated in Fig. \ref{fig:311framework}, residents can submit service requests to the city's 311 service center by calling 311 or using a mobile app or online portal, based on their observation of real-world issues.
Once a request is submitted, it is routed to the appropriate city department for processing, depending on the type and description of the issue.
Table \ref{table:departmentassignedtypes} presents example city departments and the types of service requests assigned to them in a U.S. city, based on the 311 data used in this study.
For instance, the Economic and Community Development Department handles reports of illegal dumping and housing violations, while the Solid Waste Division of Public Works Department manages requests related to missed garbage or brush collection. Each department receives and queues the assigned requests, and schedules city resources accordingly. For instance, garbage collection trucks are dispatched to resolve missed pickup requests.

Each service request contains the essential information needed by the responsible city government department to understand the issue and allocate the appropriate resources. In this work, we use a publicly available 311 dataset collected from a U.S. city, containing nearly 170,000 municipal service requests with complete service time information from 2022 to 2024. The city has a population of more than 187,000 residents. An example service request is shown in Table \ref{table:requestexample}, which includes the time the request is created and completed, the responsible department, the request type, the GPS location, and the textual description provided by the resident.

\begin{table}
\centering
\caption{Attributes in each service request}
\label{table:requestexample}
\small
\begin{tabular}{ccc} 
\hline\hline
Created Date     & Completed At                & Department              \\ 
\hline
1/2/2023 8:13    & 1/4/2023 10:31              & PW - Solid Waste        \\ 
\hline\hline
Request Type     & GPS \& Council District~ & Description             \\ 
\hline
Brush Collection & (-85.2846, 35.0979) \& 2    & Holiday tree pickup  \\
\hline
\end{tabular}
\end{table}

\subsection{Challenges}
\label{sec:challenges}
The main task of this work is to estimate the service time, i.e., the duration required to complete a submitted service request, based on the information provided by the resident.
However, this task presents two distinct challenges compared to the typical delivery time estimation problem.
\begin{figure}[h]
    \centering
    \begin{minipage}[b]{0.49\linewidth}
        \centering
            \includegraphics[width=1.5in]{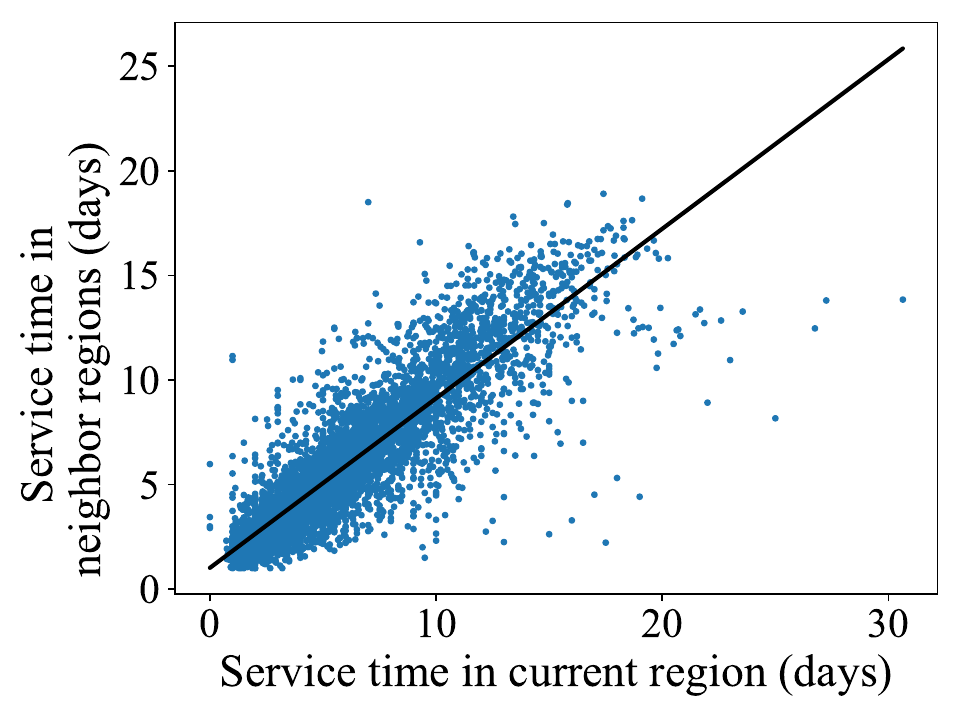}
            \caption{Spatial correlation}
        \label{fig:spatialcorrelation}
    \end{minipage}
    \hfill
    \begin{minipage}[b]{0.49\linewidth}
        \centering
        \includegraphics[width=1.5in]{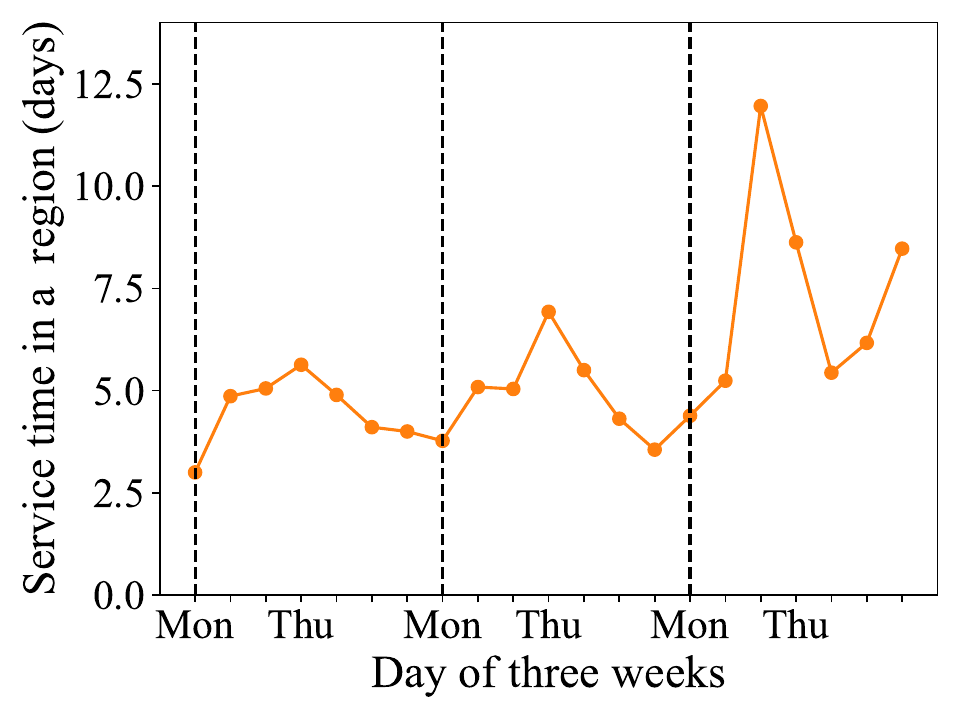}
        \caption{Temporal correlation}
        \label{fig:temporalcorrelation}
    \end{minipage}
   \vspace{-0.1in}
\end{figure}

\subsubsection{Service time correlation in spatial-temporal dimensions}
The service time of requests for a given type is not independent across spatial and temporal dimensions. Fig.~\ref{fig:spatialcorrelation} illustrates the spatial correlation of service times for a specific request type, namely brush collection. Each dot in the figure represents a region-level data point: the average service time of brush collection requests submitted in a specific region on a given day (x-axis) versus the average service time of brush collection requests submitted in its neighboring regions on the same day (y-axis). 
Fig.~\ref{fig:spatialcorrelation} highlights the spatial dependency of service times across adjacent regions, showing a positive correlation between the service time in a region and that in its neighboring regions.
Fig.~\ref{fig:temporalcorrelation} shows the temporal correlation of service time for a specific request type within a randomly selected region. Specifically, we plot the daily service time of brush collection requests submitted in the same region over a three-week period. A consistent pattern is observed across all three weeks: the service time tends to increase during the early part of the week and then decrease toward the end.
The service time on a given day is influenced by the service times in the preceding days. For example, the service times on Saturday and Sunday of the third week are higher than those of the first and second weeks, which can be attributed to elevated service times at the beginning of the third week.

   \begin{figure}[h]
    \centering
    \includegraphics[width=3.3in]{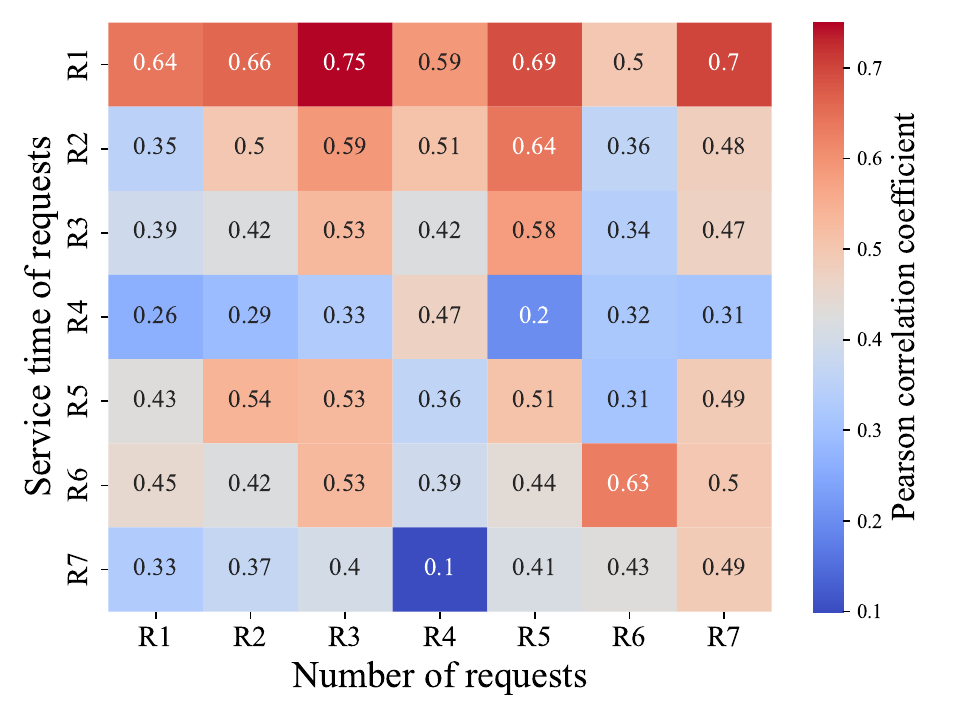}
       \vspace{-0.2in}
  \caption{Pearson correlation coefficient between request demand and  service time of different types of requests (R1: Bagged Yard Waste, R2: Brush Collection, R3: Bulk Trash, R4: Garbage container repair/replace, R5: Missed Garbage, R6: Missed Recycle, R7: New Garbage Container).}
   \label{fig:challenge1}
    \vspace{-0.1in}
\end{figure}

\subsubsection{Interaction among requests across heterogeneous types} 
As shown in Table \ref{table:departmentassignedtypes}, each department or division in the city handles several heterogeneous types of service requests.
These requests may compete for the department’s limited resources, potentially affecting the service time.
To validate the hypothesis, we compute the Pearson correlation coefficient~\cite{cohen2009pearson} between the service time of each request type and the number of submitted requests for each type.
We select seven request types assigned to the Solid Waste Division of the Department of Public Works for analysis, and the results are presented in Fig. \ref{fig:challenge1}.
We observe that the service time for a given request type is correlated with the number of other request types. For example, the Pearson correlation coefficient between the number of requests for R2 through R7 and the service time of R1 exceeds 0.5.
Therefore, it is essential to account for the potential correlation among heterogeneous request types when predicting the response time of a specific type.
    
    
    \begin{figure}[]
    \centering
    \includegraphics[width=3.0in]{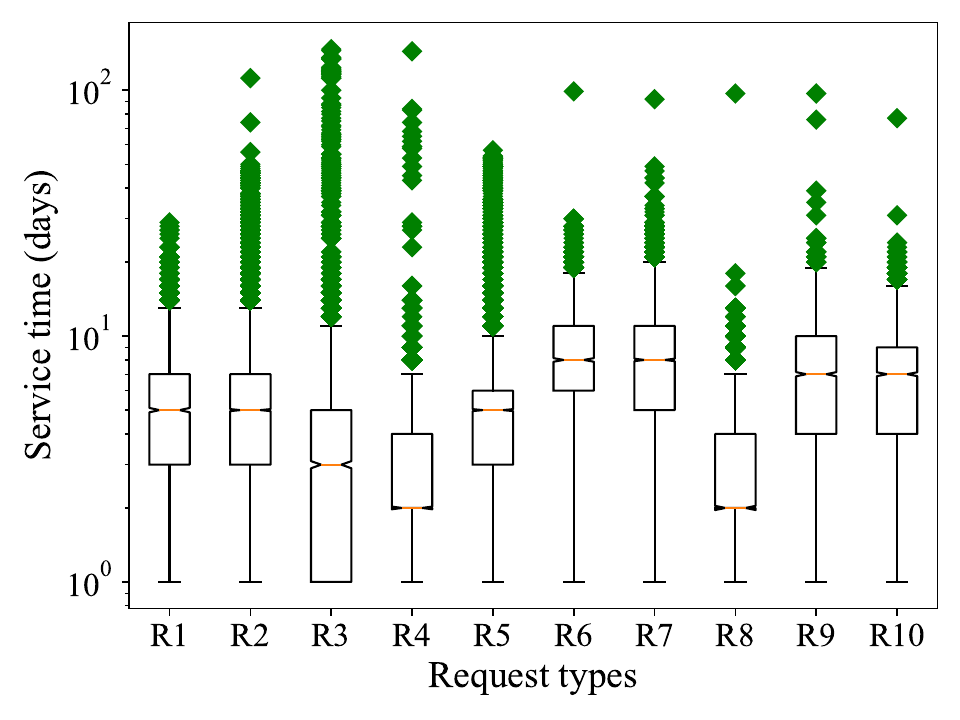}
    \vspace{-0.2in}
    \caption{Service time variance in homogeneous request types (R1: Loose Leaf Collection, R2: Bulk Trash, R3: Potholes, R4: Missed Garbage, R5: Brush Collection, R6: Recycle Container Repair, R7: Garbage container repair, R8: Missed Recycle, R9: New Garbage Container, R10: New Recycle Container)}
\label{fig:challenge2}
    \vspace{-0.2in}
\end{figure}
 
 \subsubsection{Service time variation in each request type} 
For a given department that handles heterogeneous types of service requests, even the service times of homogeneous requests exhibit significant variation, indicating the complexity of service time prediction. We select the top 10 request types based on their frequency and present the service time distribution for each type in Fig.~\ref{fig:challenge2}.
We can observe the variation of service time for all the request types. 
For instance, for R6 (missed recycle) and R7 (garbage container repair), the bottom 25\% of requests are completed in less than 6 and 5 days, respectively, while the top 25\% take more than 10 days. This indicates that one-quarter of residents experience nearly twice the service time compared to another quarter, highlighting substantial variation in service delivery.
This emphasizes the need to account for the unique characteristics of each request and the variation in service quality, even within the same type, when inferring service time.

\section{Problem Definition}


In cities with municipal service systems such as 311, we divide the area into $M$ regions. 
In this work, the regions are determined according to council districts, as indicated in the dataset and illustrated in Table~\ref{table:requestexample}.
Suppose residents submit $N$ types of service requests, with examples listed in Table~\ref{table:departmentassignedtypes}.

Each submitted request includes information such as the creation date, responsible department, request type, GPS location, council district, and a text-based description. Once the request is completed, the completion date is also recorded. 
{Below is the formal problem definition.}

\begin{definition}[Service Time Prediction for Municipal Service Requests]\label{def:problem}
\textit{Given historical service request data including the spatial, temporal, and type information, we aim to predict  the service time for a new request based on its creation date, responsible department, request type, GPS location, council district, and text-based description.
}
\end{definition}

\section{Method}

\begin{figure*}
    \centering
    \includegraphics[width=5.0in]{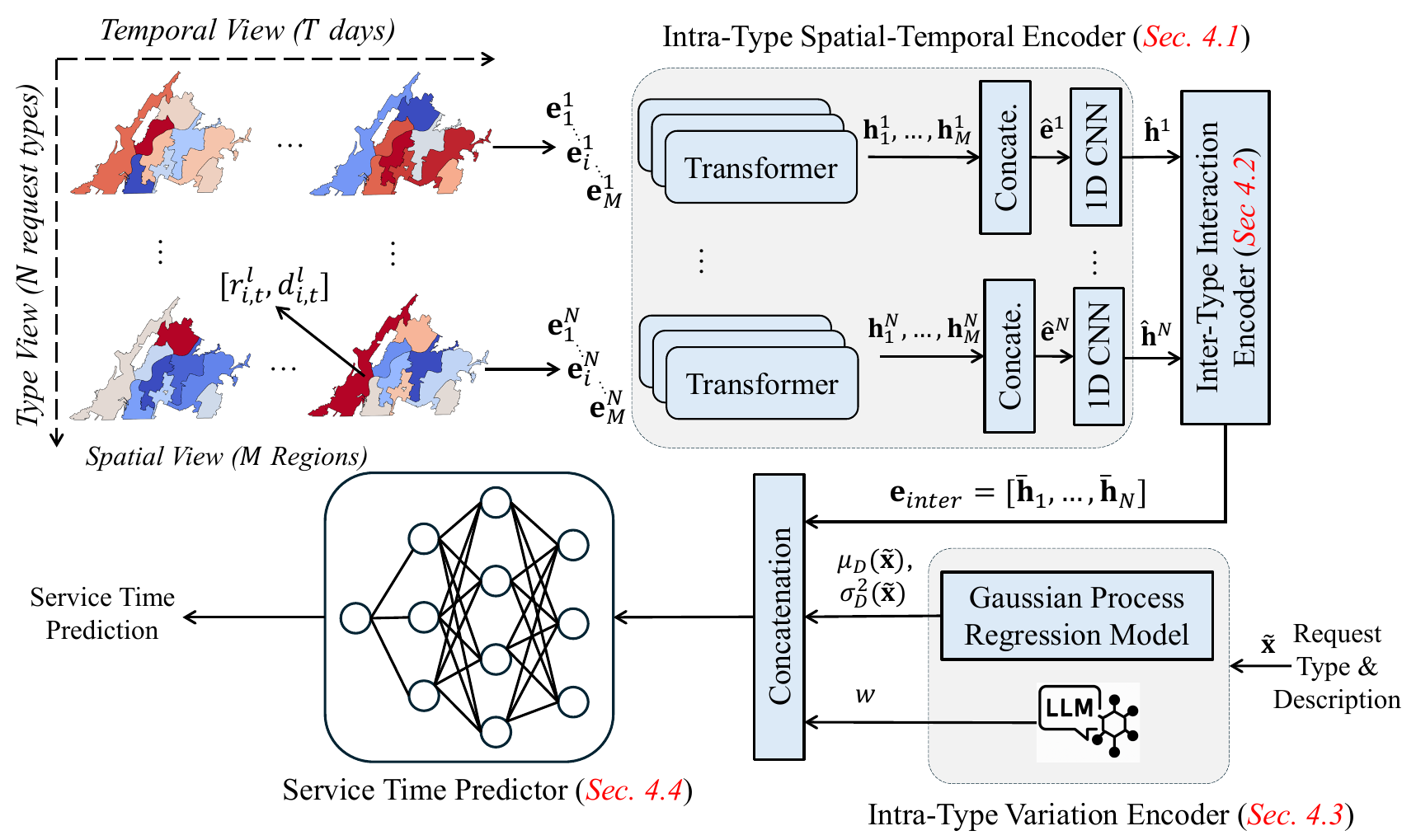}
     \vspace{-0.1in}
    \caption{MuST$^2$-Learn Framework}
     \vspace{-0.1in}
    \label{fig:modelframework}
\end{figure*}

We propose a multi-view spatial-temporal-type learning framework for service time prediction, as shown in Fig.~\ref{fig:modelframework}. The framework consists of the following components:
(i) an intra-type spatial-temporal encoder that captures spatial and temporal correlations within each request type;
(ii) an inter-type interaction encoder that models how features from heterogeneous request types influence the service time of a specific type;
(iii) an intra-type variation encoder that captures service time variations within the same request type by incorporating historical data and the unique features of a new request derived from its text-based description;
(iv) a service time predictor that estimates it for a new request based on the outputs from the inter-type interaction and intra-type variation encoders.

\subsection{Intra-Type Spatial-temporal Encoder}


To model the spatio-temporal dependencies for each request type $l \in \mathcal{L}$, where $\mathcal{L}$ denotes the set of all request types, we propose a novel \emph{Intra-type Spatio-temporal Encoder}. Specifically, it captures the temporal correlations with temporal self-attention and captures the spatial correlations with a 1D convolutional neural network (CNN) in a hierarchical architecture. 

\noindent 
\textbf{Temporal Correlation:} 
We first employ the Transformer model~\cite{vaswani2017attention} to capture the temporal dynamics of service demand patterns. For each region $i \in \mathcal{M}$ (where $\mathcal{M}$ is the set of regions) and request type $l$, the model processes a historical sequence of type-region embeddings:

\begin{equation}
    \mathbf{e}_{i}^l = [\mathbf{e}_{i,t}^l]_{t=t'-T}^{t'-1} \in \mathbb{R}^{2T},
\end{equation}
where $t'$ represents the current day and $T$ is the temporal look-back window. Each embedding vector $\mathbf{e}_{i,t}^l$ consists of two key features:

\begin{equation}
    \mathbf{e}_{i,t}^l = [r^l_{i,t}, d^l_{i,t}],
\end{equation}
where:
\begin{itemize}[leftmargin=*]
    \item $r^l_{i,t} \in \mathbb{R}$ is the request volume for type $l$ in region $i$ at day $t$.
    \item $d^l_{i,t} \in \mathbb{R}$ represents the average service time for these requests.
\end{itemize}

The Transformer maps the input sequence $\mathbf{e}_{i}^l$ to a latent temporal representation:

\begin{equation}
    \mathbf{h}_{i}^l = \text{Transformer}(\mathbf{e}_{i}^l),
\end{equation}
capturing long-range dependencies through its self-attention mechanism. This process generates $M \times N$ hidden states, where $M = |\mathcal{M}|$ is the number of regions and $N = |\mathcal{L}|$ is the number of request types.




\noindent
\textbf{Spatial Correlation:} 
To further capture spatial correlations between regions (e.g., request demand propagation effects, the service time of a given request type in one region influences neighboring regions), we design a 1D-CNN that operates on the Transformer outputs:

\begin{equation}
    \hat{\mathbf{e}}^l = [\mathbf{h}^l_i]_{i=1}^M \in \mathbb{R}^{M \times d},
\end{equation}
where $d$ is the dimension of the hidden states. we adopt 1D-CNN because the irregular spatial layout of city districts (Fig.~\ref{fig:modelframework}) makes traditional 2D convolutions suboptimal. 

\subsection{Inter-type Interaction Encoder}

While the intra-type spatial-temporal encoders capture spatial and temporal correlations within each request type, it is also essential to model the inter-type interaction across heterogeneous request types, as illustrated in Fig.~\ref{fig:challenge1}. The motivation is that city departments operate with limited resources shared among different types of service requests. 
An increased demand for one request type may reduce the availability of resources for others, potentially leading to longer service times. To capture this dynamic, we develop an attention-based inter-type interaction encoder that learns how the service demand and service time of one type influence those of other types. 

Let $\Bar{\mathbf{e}} = [\hat{\mathbf{h}}_1; \hat{\mathbf{h}}_2; \ldots; \hat{\mathbf{h}}_N]$ denote the embeddings obtained from the intra-type spatial-temporal encoders for all $N$ request types. The inter-type interaction encoder assesses the relevance of each type's embedding in $\Bar{\mathbf{e}}$ to the service time prediction of a given type $l$. Specifically, a higher attention weight is assigned to $\hat{\mathbf{h}}_{l'}$ if the request type $l'$ has a stronger influence on type $l$. The resulting output from the inter-type encoder is represented as $\mathbf{e}_{inter}=[\Bar{\mathbf{h}}_1; \Bar{\mathbf{h}}_2; \ldots; \Bar{\mathbf{h}}_N]$, which serves as part of the input to the final service time prediction module.

\subsection{Intra-Type Variation Encoder}
As shown in Fig.~\ref{fig:challenge2}, significant variation in service time still exists within each individual request type, even though the inter-type encoder captures correlations among heterogeneous request types. 
To address this, we introduce an intra-type variation encoder that models the service time variation among requests of the same type. This encoder leverages historical service data and the text-based descriptions of the requests. 

\noindent \textbf{Intra-type Variation Modeling:} First, we employ Gaussian Process Regression (GPR)~\cite{shah2025interference,8317796,Roberts2013Gaussian} to quantify and model the inherent uncertainty in service time prediction.
GPR offers a non-parametric Bayesian framework, providing both predictive mean and variances, which in turn helps to capture how confident the model is in its prediction.
We use $D=(\mathbf{X},\mathbf{y})$ to represent the training dataset, where 
$\mathbf{X} \in \mathbb{R}^{L\times F}$ denotes the training input with $L$ training samples, each with $F$ features, and $ \mathbf{y} \in \mathbb{R}^L$ is the corresponding vector of service times.
We assume a Gaussian process prior is given by:
\begin{equation}
f(\mathbf{x}) \sim \mathcal{GP}(\mu(\mathbf{x}), k(\mathbf{x}, \mathbf{x}')).
\end{equation}
Here, $k(\mathbf{x}, \mathbf{x}')$ is a kernel function, more specifically, a radial basis function~\cite{Arora2023review} in this work. With training samples, the posterior of $f(\mathbf{x})$ is used to make predictions at new points $\mathbf{x_*}$. According to the definition of GPR, these predictions of new points will also follow a Gaussian distribution: 
\begin{align}
\mu_D(\mathbf{x}_*) &= \mathbf{k}_*^\top (\mathbf{K} + \alpha \mathbf{I})^{-1} \mathbf{y} \\
\sigma_D^2(\mathbf{x}_*) &= k(\mathbf{x}_*, \mathbf{x}_*) - \mathbf{k}_*^\top (\mathbf{K} + \alpha \mathbf{I})^{-1} \mathbf{k}_*.
\end{align}
Here, $\mathbf{k}_* = [k(\mathbf{x}_1, \mathbf{x}_*), \ldots, k(\mathbf{x}_L, \mathbf{x}_*)]^\top$ is the covariance vector between the training inputs $D = [\mathbf{x}_1, ..., \mathbf{x}_L]$ and the test input  $\mathbf{x_*}$, $\mathbf{K} \in \mathbb{R}^{L \times L}$ is the covariance matrix over the training inputs, $\mathbf{I}$ is the identity matrix, and $\alpha$ is the noise term accounting for observation noise.\par

For each service request type, we build a training dataset from its historical records and corresponding service times, which is then used to train a GPR model. Applying the procedure to all $N$ request types yields $N$ distinct GPR models.
After training, given a newly submitted service request $\mathbf{\Tilde{x}}$, we use the GPR model corresponding to its request type to compute the predictive mean and standard deviation:


\begin{align}
&\Tilde{y}  \sim \mathcal{N} \big({\mu}(\mathbf{\Tilde{x}}), {\sigma}^2(\mathbf{\Tilde{x}}) \big),
\\
&{\mu} (\mathbf{\Tilde{x}}) = \mathbb{E}[\Tilde{y}]= {\mu}_{D} (\mathbf{\Tilde{x}}),
\\
& {\sigma}^2 (\mathbf{\Tilde{x}}) = \text{Var}[\Tilde{y}] ={\sigma}^2_{D} (\mathbf{\Tilde{x}}), 
\end{align}
where ${\mu}_{D}(\Tilde{x})$ and ${\sigma}^2_{D}(\Tilde{x})$ denote the predictive mean and variance from the GPR model trained on $D$. 
These predictions reflect the variation in service time within the specific request type and will be incorporated into the intra-type embedding.

\begin{table}[t]
\centering
\caption{Examples of request description and workload}
\label{table:workload}
\begin{tabular}{l|l|l} 
\hline
Text description                                                                                      & Request type                                               & Workload  \\ 
\hline
\begin{tabular}[c]{@{}l@{}}2 mattress, bed frame, 3 desk, \\2 cabinet, hamper, flag pole\end{tabular} & Bulk trash                                                 & 8         \\ 
\hline
2 mattress                                                                                            & Bulk trash                                                 & 6         \\ 
\hline
sofa pick up in rear                                                                                  & Bulk trash                                                 & 4         \\ 
\hline
2 chairs                                                                                              & Bulk trash                                                 & 3         \\ 
\hline
\begin{tabular}[c]{@{}l@{}}13 bags of leaves and then \\a Christmas tree\end{tabular}                 & \begin{tabular}[c]{@{}l@{}}Bagged yard\\waste\end{tabular} & 7         \\ 
\hline
9 black bags of
  yard debris.                                                                        & \begin{tabular}[c]{@{}l@{}}Bagged yard\\waste\end{tabular} & 5         \\ 
\hline
2 bags of leaves                                                                                      & \begin{tabular}[c]{@{}l@{}}Bagged yard\\waste\end{tabular} & 3         \\
\hline
\end{tabular}
\end{table}

\begin{figure}[t]
    \centering
    \includegraphics[width=2.7in]{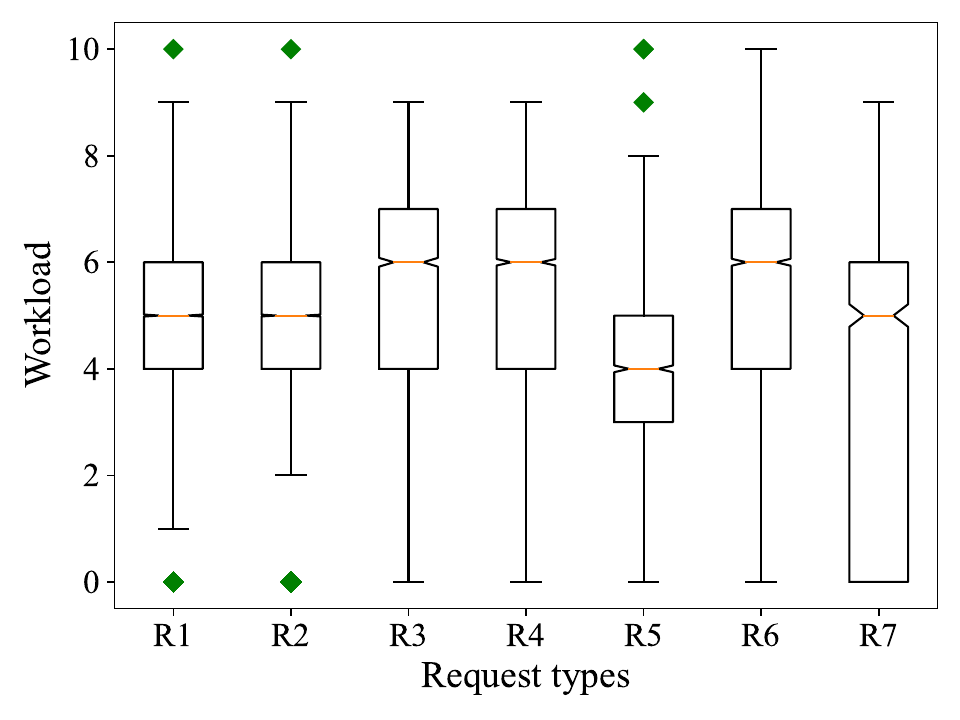}
     \vspace{-0.1in}
    \caption{Distribution of request workload (R1: Bagged Yard Waste, R2: Brush Collection, R3: Bulk Trash, R4: Garbage container repair/replace, R5: Missed Garbage, R6: Missed Recycle, R7: New Garbage Container).}
    \vspace{-0.1in}
    \label{fig:workloaddistribution}
\end{figure}

\noindent \textbf{LLM-based Request Workload Assignment:} Second, intuitively, the text-based service request description can offer useful information about the expected workload required to address a specific request, which may influence the service time. 
To extract the valuable information, we use large language models (LLMs)~\cite{matallm} to assess the workload associated with each request based on its textual description.
In detail, each service request is assigned a workload index $w\in [0,10]$ based on its text description, indicating the estimated effort required to address it. A value of $w=1$ corresponds to a low workload, while $w=10$ indicates a high workload.
For requests with missing text descriptions, we assign  $w=0$.
Table \ref{table:workload} provides examples of service request text descriptions along with the corresponding workload levels assigned by the LLM. Based on the specific content of each description, a workload score is assigned to reflect the estimated effort required. Fig. \ref{fig:workloaddistribution} illustrates the distribution of workload scores across different request types. Notably, within each type, there exist requests with significant deviations from the mean workload. 
For instance, while 50\% of the requests related to bagged yard waste have workload scores between 4 and 6, a portion of them receive higher scores, such as 8 or 9.

In summary, the embedding of a newly submitted request, derived from the output of the intra-type encoder, is constructed by concatenating the predicted mean and variance for the corresponding request type with the workload assigned based on the request description. This embedding is denoted as  $\mathbf{e}_{intra}=[{\mu}_{D}(\Tilde{\textbf{x}}), {\sigma}^2_{D}(\Tilde{\textbf{x}}), w]$.

\subsection{Service Time Predictor}

At last, we input the learned embedding, denoted as $\mathbf{e}=[\mathbf{e}_{inter}, \mathbf{e}_{intra}]$, which combines outputs from the inter-type and intra-type encoders, into a service time predictor. 
Specifically, we implement the predictor with an efficient Multilayer Perceptron (MLP) to perform regression for estimating the service time of a newly submitted request $\mathbf{\Tilde{x}}$.
The predictor can be described by the following mathematical equations:
\begin{align}
    \mathbf{h}_0 & = \mathbf{e},  \\
    \mathbf{h}_j &= \psi (\mathbf{W}_j\mathbf{h}_{j-1}+\mathbf{b}_j), \ \ \  j\in [1,n] \\
    \hat{y} & = \sigma (\mathbf{h}_{n}), 
\end{align}
where $n$ is the number of hidden layers in MLP, and $\mathbf{W}_j$ and $\mathbf{b}_j$ are the weight matrix and bias vector of layer $j$.
$\psi(\cdot)$ is the activation function of the fully connected layers, which is \textit{ReLU} in this work.
$\sigma(\cdot)$ is the activation function used to output the estimated service time, i.e., $\Tilde{y}$, for a newly submitted request $\Tilde{\textbf{x}}$.
The service time prediction is a regression problem, and we use the mean squared error as the metric of our loss function to train the proposed model.

\section{Evaluation}
In this section, we comprehensively evaluate our proposed method MuST$^2$-learn for predicting service time of newly submitted 311 requests. 

\begin{table*}[]
\caption{Performance comparison across six request types using the Chattanooga dataset}
\label{table:overallperformance}
\begin{tabular}{ccccccccccccc}
\hline
\multicolumn{1}{c|}{}            & \multicolumn{4}{c|}{New Garbage Container}                                                                                                           & \multicolumn{4}{c|}{Bulk Trash}                                                                                                                      & \multicolumn{4}{c}{Missed Recycle}                                                                                                  \\ \hline
\multicolumn{1}{c|}{Methods}     & \multicolumn{1}{c|}{MAE}           & \multicolumn{1}{c|}{MSE}           & \multicolumn{1}{c|}{RMSE}          & \multicolumn{1}{c|}{MAPE}             & \multicolumn{1}{c|}{MAE}           & \multicolumn{1}{c|}{MSE}           & \multicolumn{1}{c|}{RMSE}          & \multicolumn{1}{c|}{MAPE}             & \multicolumn{1}{c|}{MAE}           & \multicolumn{1}{c|}{MSE}           & \multicolumn{1}{c|}{RMSE}          & MAPE                 \\ \hline
\multicolumn{1}{c|}{ARIMA}       & \multicolumn{1}{c|}{3.29}          & \multicolumn{1}{c|}{14.75}         & \multicolumn{1}{c|}{3.51}          & \multicolumn{1}{c|}{97.00\%}          & \multicolumn{1}{c|}{3.99}          & \multicolumn{1}{c|}{21.20}         & \multicolumn{1}{c|}{4.50}          & \multicolumn{1}{c|}{98.69\%}          & \multicolumn{1}{c|}{1.35}          & \multicolumn{1}{c|}{2.58}          & \multicolumn{1}{c|}{1.59}          & 48.93\%              \\ \hline
\multicolumn{1}{c|}{Prophet}     & \multicolumn{1}{c|}{1.40}          & \multicolumn{1}{c|}{4.35}          & \multicolumn{1}{c|}{1.50}          & \multicolumn{1}{c|}{27.09\%}          & \multicolumn{1}{c|}{1.69}          & \multicolumn{1}{c|}{27.69}         & \multicolumn{1}{c|}{3.54}          & \multicolumn{1}{c|}{67.84\%}          & \multicolumn{1}{c|}{1.38}          & \multicolumn{1}{c|}{3.55}          & \multicolumn{1}{c|}{1.46}          & 66.08\%              \\ \hline
\multicolumn{1}{c|}{Transformer} & \multicolumn{1}{c|}{3.77}          & \multicolumn{1}{c|}{25.17}         & \multicolumn{1}{c|}{5.00}          & \multicolumn{1}{c|}{83.69\%}          & \multicolumn{1}{c|}{1.66}          & \multicolumn{1}{c|}{4.24}          & \multicolumn{1}{c|}{2.06}          & \multicolumn{1}{c|}{54.59\%}          & \multicolumn{1}{c|}{2.63}          & \multicolumn{1}{c|}{9.53}          & \multicolumn{1}{c|}{3.07}          & 101.69\%             \\ \hline
\multicolumn{1}{c|}{CNN}         & \multicolumn{1}{c|}{3.02}          & \multicolumn{1}{c|}{13.77}         & \multicolumn{1}{c|}{3.71}          & \multicolumn{1}{c|}{154.23\%}         & \multicolumn{1}{c|}{8.32}          & \multicolumn{1}{c|}{91.97}         & \multicolumn{1}{c|}{9.59}          & \multicolumn{1}{c|}{330.16\%}         & \multicolumn{1}{c|}{2.79}          & \multicolumn{1}{c|}{7.78}          & \multicolumn{1}{c|}{2.79}          & 205.48\%             \\ \hline
\multicolumn{1}{c|}{GPR}         & \multicolumn{1}{c|}{4.52}          & \multicolumn{1}{c|}{29.38}         & \multicolumn{1}{c|}{5.42}          & \multicolumn{1}{c|}{234.94\%}         & \multicolumn{1}{c|}{2.17}          & \multicolumn{1}{c|}{11.27}         & \multicolumn{1}{c|}{3.36}          & \multicolumn{1}{c|}{60.01\%}          & \multicolumn{1}{c|}{1.96}          & \multicolumn{1}{c|}{4.51}          & \multicolumn{1}{c|}{2.12}          & 140.06\%             \\ \hline
\multicolumn{1}{c|}{DeepSTA}     & \multicolumn{1}{c|}{3.54}          & \multicolumn{1}{c|}{18.13}         & \multicolumn{1}{c|}{3.99}          & \multicolumn{1}{c|}{63.83\%}          & \multicolumn{1}{c|}{2.07}          & \multicolumn{1}{c|}{6.63}          & \multicolumn{1}{c|}{2.52}          & \multicolumn{1}{c|}{62.04\%}          & \multicolumn{1}{c|}{1.62}          & \multicolumn{1}{c|}{2.87}          & \multicolumn{1}{c|}{1.65}          & 106.19\%             \\ \hline
\multicolumn{1}{c|}{IGTrans}     & \multicolumn{1}{c|}{3.68}          & \multicolumn{1}{c|}{20.27}         & \multicolumn{1}{c|}{4.48}          & \multicolumn{1}{c|}{121.62\%}         & \multicolumn{1}{c|}{2.47}          & \multicolumn{1}{c|}{9.82}          & \multicolumn{1}{c|}{3.13}          & \multicolumn{1}{c|}{80.38\%}          & \multicolumn{1}{c|}{2.41}          & \multicolumn{1}{c|}{9.84}          & \multicolumn{1}{c|}{3.05}          & 128.24\%             \\ \hline
\multicolumn{1}{c|}{\textbf{Our MuST$^2$-Learn}} & \multicolumn{1}{c|}{\textbf{0.21}} & \multicolumn{1}{c|}{\textbf{0.96}}          & \multicolumn{1}{c|}{\textbf{0.98}} & \multicolumn{1}{c|}{\textbf{19.95\%}} & \multicolumn{1}{c|}{\textbf{0.81}} & \multicolumn{1}{c|}{\textbf{3.78}} & \multicolumn{1}{c|}{\textbf{1.94}} & \multicolumn{1}{c|}{\textbf{24.90\%}} & \multicolumn{1}{c|}{\textbf{0.13}} & \multicolumn{1}{c|}{\textbf{0.25}} & \multicolumn{1}{c|}{\textbf{0.50}} & \textbf{16.40\%}     \\ \hline
\multicolumn{1}{l}{}             & \multicolumn{1}{l}{}               & \multicolumn{1}{l}{}               & \multicolumn{1}{l}{}               & \multicolumn{1}{l}{}                  & \multicolumn{1}{l}{}               & \multicolumn{1}{l}{}               & \multicolumn{1}{l}{}               & \multicolumn{1}{l}{}                  & \multicolumn{1}{l}{}               & \multicolumn{1}{l}{}               & \multicolumn{1}{l}{}               & \multicolumn{1}{l}{} \\ \hline
\multicolumn{1}{c|}{}            & \multicolumn{4}{c|}{Missed Garbage}                                                                                                                  & \multicolumn{4}{c|}{Brush Collection}                                                                                                                & \multicolumn{4}{c}{Bagged Yard Waste}                                                                                               \\ \hline
\multicolumn{1}{c|}{Methods}     & \multicolumn{1}{c|}{MAE}           & \multicolumn{1}{c|}{MSE}           & \multicolumn{1}{c|}{RMSE}          & \multicolumn{1}{c|}{MAPE}             & \multicolumn{1}{c|}{MAE}           & \multicolumn{1}{c|}{MSE}           & \multicolumn{1}{c|}{RMSE}          & \multicolumn{1}{c|}{MAPE}             & \multicolumn{1}{c|}{MAE}           & \multicolumn{1}{c|}{MSE}           & \multicolumn{1}{c|}{RMSE}          & MAPE                 \\ \hline
\multicolumn{1}{c|}{ARIMA}       & \multicolumn{1}{c|}{1.07}          & \multicolumn{1}{c|}{1.78}          & \multicolumn{1}{c|}{1.25}          & \multicolumn{1}{c|}{50.56\%}          & \multicolumn{1}{c|}{2.35}          & \multicolumn{1}{c|}{9.50}          & \multicolumn{1}{c|}{3.04}          & \multicolumn{1}{c|}{50.82\%}          & \multicolumn{1}{c|}{1.67}          & \multicolumn{1}{c|}{5.22}          & \multicolumn{1}{c|}{2.21}          & 70.34\%              \\ \hline
\multicolumn{1}{c|}{Prophet}     & \multicolumn{1}{c|}{0.68}          & \multicolumn{1}{c|}{1.09}          & \multicolumn{1}{c|}{0.72}          & \multicolumn{1}{c|}{40.75\%}          & \multicolumn{1}{c|}{1.27}          & \multicolumn{1}{c|}{5.39}          & \multicolumn{1}{c|}{1.58}          & \multicolumn{1}{c|}{25.94\%}          & \multicolumn{1}{c|}{1.52}          & \multicolumn{1}{c|}{4.28}          & \multicolumn{1}{c|}{1.75}          & 31.22\%              \\ \hline
\multicolumn{1}{c|}{Transformer} & \multicolumn{1}{c|}{1.07}          & \multicolumn{1}{c|}{1.78}          & \multicolumn{1}{c|}{1.33}          & \multicolumn{1}{c|}{57.16\%}          & \multicolumn{1}{c|}{1.94}          & \multicolumn{1}{c|}{5.28}          & \multicolumn{1}{c|}{2.29}          & \multicolumn{1}{c|}{37.94\%}          & \multicolumn{1}{c|}{2.13}          & \multicolumn{1}{c|}{7.87}          & \multicolumn{1}{c|}{2.80}          & 43.98\%              \\ \hline
\multicolumn{1}{c|}{CNN}         & \multicolumn{1}{c|}{1.96}          & \multicolumn{1}{c|}{4.43}          & \multicolumn{1}{c|}{2.11}          & \multicolumn{1}{c|}{160.68\%}         & \multicolumn{1}{c|}{0.80}          & \multicolumn{1}{c|}{\textbf{0.77}} & \multicolumn{1}{c|}{\textbf{0.88}} & \multicolumn{1}{c|}{\textbf{20.09\%}} & \multicolumn{1}{c|}{4.38}          & \multicolumn{1}{c|}{24.23}         & \multicolumn{1}{c|}{4.92}          & 62.77\%              \\ \hline
\multicolumn{1}{c|}{GPR}         & \multicolumn{1}{c|}{1.12}          & \multicolumn{1}{c|}{2.27}          & \multicolumn{1}{c|}{1.51}          & \multicolumn{1}{c|}{54.13\%}          & \multicolumn{1}{c|}{2.12}          & \multicolumn{1}{c|}{9.36}          & \multicolumn{1}{c|}{3.06}          & \multicolumn{1}{c|}{63.53\%}          & \multicolumn{1}{c|}{1.80}          & \multicolumn{1}{c|}{6.01}          & \multicolumn{1}{c|}{2.45}          & 53.53\%              \\ \hline
\multicolumn{1}{c|}{DeepSTA}     & \multicolumn{1}{c|}{1.10}          & \multicolumn{1}{c|}{1.67}          & \multicolumn{1}{c|}{1.76}          & \multicolumn{1}{c|}{90.57\%}          & \multicolumn{1}{c|}{1.05}          & \multicolumn{1}{c|}{1.43}          & \multicolumn{1}{c|}{1.05}          & \multicolumn{1}{c|}{26.13\%}          & \multicolumn{1}{c|}{5.54}          & \multicolumn{1}{c|}{35.01}         & \multicolumn{1}{c|}{5.87}          & 65.61\%              \\ \hline
\multicolumn{1}{c|}{IGTrans}     & \multicolumn{1}{c|}{2.41}          & \multicolumn{1}{c|}{9.84}          & \multicolumn{1}{c|}{3.05}          & \multicolumn{1}{c|}{128.24\%}         & \multicolumn{1}{c|}{3.17}          & \multicolumn{1}{c|}{18.40}         & \multicolumn{1}{c|}{4.18}          & \multicolumn{1}{c|}{75.89\%}          & \multicolumn{1}{c|}{3.11}          & \multicolumn{1}{c|}{15.84}         & \multicolumn{1}{c|}{3.80}          & 79.66\%              \\ \hline
\multicolumn{1}{c|}{\textbf{Our MuST$^2$-Learn}} & \multicolumn{1}{c|}{\textbf{0.13}} & \multicolumn{1}{c|}{\textbf{0.14}} & \multicolumn{1}{c|}{\textbf{0.37}} & \multicolumn{1}{c|}{\textbf{16.24\%}} & \multicolumn{1}{c|}{\textbf{0.54}} & \multicolumn{1}{c|}{2.09}          & \multicolumn{1}{c|}{1.45}          & \multicolumn{1}{c|}{23.19\%}          & \multicolumn{1}{c|}{\textbf{0.16}} & \multicolumn{1}{c|}{\textbf{0.63}} & \multicolumn{1}{c|}{\textbf{0.79}} & \textbf{12.68\%}     \\ \hline
\end{tabular}
\end{table*}

\subsection{Evaluation Methodology}

\textbf{Datasets:} We use the 311 service request data collected from the City of Chattanooga, Tennessee, which contains nearly 170,000 municipal service requests with complete service time information from 2022 to 2024.
The key attributes of each service request record are as follows:
\vspace{-0.05in}
\begin{itemize}[leftmargin=*]
  \item \textit{Created Date}: The timestamp indicating the submission time of the request.
  \item \textit{Completed At}: The timestamp indicating the completion time of the request.
  \item \textit{Department}: The city department responsible for resolving the request (e.g., ``PW - Solid Waste'').
  \item \textit{Request Type}: The specific type of service request (e.g., ``Brush Collection'' and ``Bulk Trash'').
  \item \textit{GPS Locations \& Council District}: It specifies the geographic coordinates of the submitted request and identifies the corresponding council district.
  \item \textit{Description}: A brief textual description of the request.
  \item \textit{Response Time}: This column is calculated from ``Created Date" and ``Completed At" column. Highest response time considered for model analysis is 80 days. \vspace{-0.05in}
\end{itemize}

To evaluate the generalizability of MuST$^2$-Learn, we assess its performance on an additional dataset collected from the City of Newark, New Jersey~\cite{newark}, and compare it with baseline methods. This dataset contains approximately 7,500 service requests, each including the attributes used previously, except for the council district. To support region-based modeling, we partition the city into regions defined by census tracts and assign each request to a tract based on its GPS coordinates.

\begin{table*}[t]
\caption{Performance comparison using the Newark dataset measured by MAPE (R1: Pothole Complaint; R2: Animal Complaint; R3: Rodent Infestation; R4: Traffic : Signal, Signage, Light Pole (an Maintenance Issues); R5: Illegal Activity (Alcohol, Gambling, Drugs, Prostitution); R6: Noise Disturbance)}
\label{table:newarkresults}

\begin{tabular}{c|c|c|c|c|c|c}
\hline
Methods     & R1               & R2              & R3              & R4              & R5              & R6              \\ \hline
ARIMA       & 83.82\%          & 102.52\%        & 97.52\%         & 140.84\%        & 95.73\%         & 87.12\%         \\ \hline
Prophet     & 85.10\%          & \textbf{3.52\%} & 37.17\%         & 41.93\%         & 13.32\%         & 58.18\%         \\ \hline
Transformer & 138.59\%         & 74.28\%         & 122.44\%        & 91.85\%         & 83.21\%         & 125.61\%        \\ \hline
CNN         & 194.96\%         & 957.78\%        & 420.82\%        & 237.31\%        & 407.68\%        & 987.79\%        \\ \hline
GPR         & 78.86\%          & 348.86\%        & 680.96\%        & 186.24\%        & 176.03\%        & 191.77\%        \\ \hline
DeepSTA     & 87.06\%          & 173.21\%        & 309.73\%        & 94.05\%         & 131.34\%        & 61.57\%         \\ \hline
IGTrans     & 110.08\%         & 311.04\%        & 494.49\%        & 201.66\%        & 126.27\%        & 166.97\%        \\ \hline
\textbf{Our MuST$^2$-Learn} & \textbf{11.89\%} & 8.31\%          & \textbf{9.17\%} & \textbf{8.56\%} & \textbf{8.23\%} & \textbf{8.31\%} \\ \hline
\end{tabular}
\end{table*}


\noindent \textbf{Baselines:}
To comprehensively evaluate the proposed method, we implement seven state-of-the-art approaches and compare the performance of MuST$^2$-Learn against them.
\begin{itemize}[leftmargin=*]
\item \textbf{ARIMA}~\cite{shumway2017arima}: It is a traditional method to forecast future values based only on the linear correlation of past values and past errors.

\item \textbf{Prophet}~\cite{wang2022community}: It is a forecasting tool, designed to handle seasonal, trend, and holiday effects in time series data. 

\item \textbf{Transformer}~\cite{vaswani2017attention}: It is an advanced deep learning architecture, which is used to model the temporal correlation.

\item \textbf{CNN}~\cite{ige2024state}: It is a deep neural network model for capturing the spatial correlation.

\item \textbf{GPR}~\cite{8317796}: It is a non-parametric, Bayesian approach to regression tasks, which outputs a probability distribution function for addressing the inherent variation.

\item \textbf{DeepSTA}~\cite{yi2023deepsta}: It is a spatial-temporal attention-based framework that is capable of capturing the correlation in spatial-temporal dimensions. 

\item \textbf{IGTrans}~\cite{zhou2023inductive}: An inductive graph transformer-based method that integrates multiplex information encoded by a graph neural network, adjusted to capture correlations among heterogeneous request types.

\end{itemize}

\noindent \textbf{Metrics:} In the experiments, we use MAE, MSE, RMSE, and MAPE as evaluation metrics to compare the performance of different methods.

\noindent \textbf{LLM:} To enrich the dataset with workload estimates derived from text-based request descriptions, we utilize the Meta LLaMA3 8B model.
Each request's textual description, along with its corresponding day, week, and year, are provided as input.
The model generates a workload score for each request, along with reasoning to justify the assigned workload.

\noindent \textbf{Settings}: Our experiments are conducted on a PC equipped with an Intel(R) Core(TM) Ultra 5 125U CPU @ 3.60GHz, 16 GB of RAM, and an NVIDIA RTX 4090 GPU.
It takes less than 1 ms to generate the predicted service time for a given request.
In the experiments, we use 80\% of the dataset for training the proposed model and the remaining 20\% for evaluating the performance of different methods.
In the experiment, the look-back window size is set to 14 days, i.e., $T = 14$. The Transformer module consists of 32 hidden units and 4 attention heads. The attention-based inter-type encoder is configured with 64 hidden units and 4 attention heads. The multi-layer perceptron (MLP) predictor contains 64 hidden units. The learning rate is 0.001.

To simplify the presentation, we refer to the request types as follows throughout the rest of the evaluation: R1 for "Bulk Trash", R2 for "Brush Collection", R3 for "Missed Garbage", R4 for "Missed Recycle", R5 for "Bagged Yard Waste", and R6 for "New Garbage Container".



\vspace{-0.1in}
\subsection{Overall Performance}

Table~\ref{table:overallperformance} presents the performance comparison using the Chattanooga dataset.
Due to space limitations, we report results for six request types with the highest service demand, which together account for nearly 60\% of all service requests.
There are several observations.
The first observation is that MuST$^2$-Learn significantly achieves the best performance across most request types, except for brush collection, where it ranks second. 
In detail, it reduces the MAE by at least 32.5\% for all request types compared to all other methods.

Second, compared to other methods that capture only a subset of spatial, temporal, or type correlations, the superior performance of MuST$^2$-Learn demonstrates the importance of jointly modeling all three dimensions. For instance, MuST$^2$-Learn reduces the MAPE by up to 84.6\% and decreases the MAE by up to 97.1\% compared to DeepSTA, which models spatial-temporal correlations using an attention mechanism.

Third, MuST$^2$-Learn does not outperform DeepSTA and CNN for brush collection requests in terms of MSE and RMSE. A possible reason is that the Chattanooga Public Works Department changed its brush pickup policy in August 2024, shifting from a fixed calendar schedule to an on-demand service. As a result, the training data contain incorrect information about the intra-type variation and inter-type interaction under the new policy.

Fourth, Prophet, a linear time series model that incorporates seasonal, trend, and holiday effects, outperforms the other deep neural network-based methods. This suggests that the service time has strong long-term temporal patterns related to seasons and holidays, which are not well captured by models using a look-back window of at most one month. This finding is also consistent with the observation in~\cite{wang2022community}.


\begin{figure}
    \centering
    \includegraphics[width=.95\linewidth]{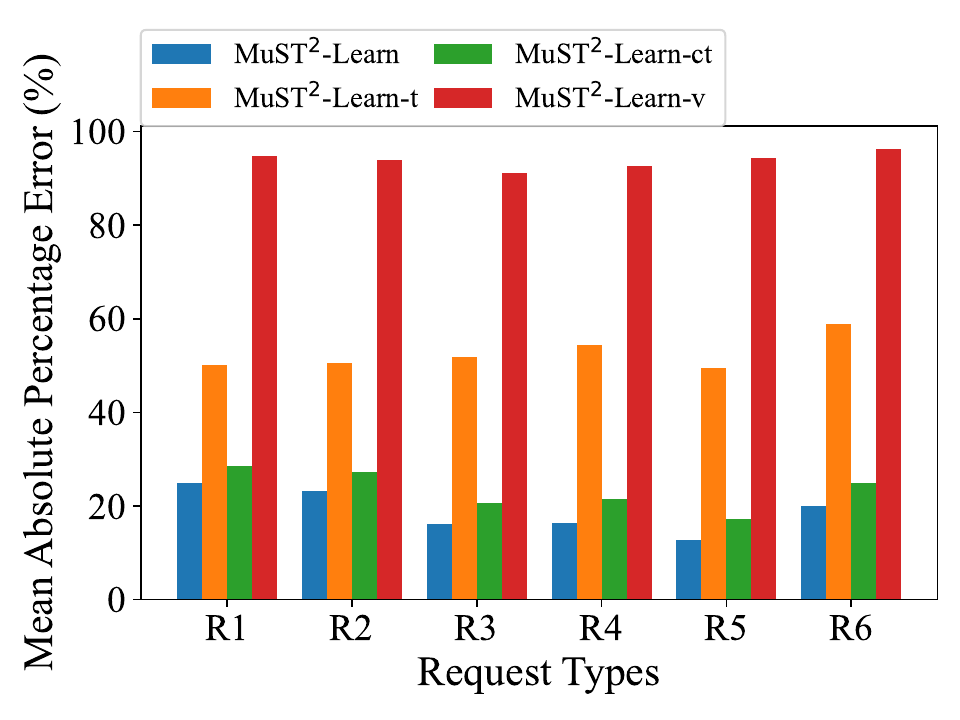}
    \vspace{-0.2in}
   \caption{Evaluation of MuST$^2$-Learn variants}
   \label{fig:ablationstudy}
   \vspace{-0.2in}
\end{figure}

\begin{figure*}[!t]
    \centering
        \begin{minipage}[b]{0.32\linewidth}
        \centering
        \includegraphics[width=2.3in]{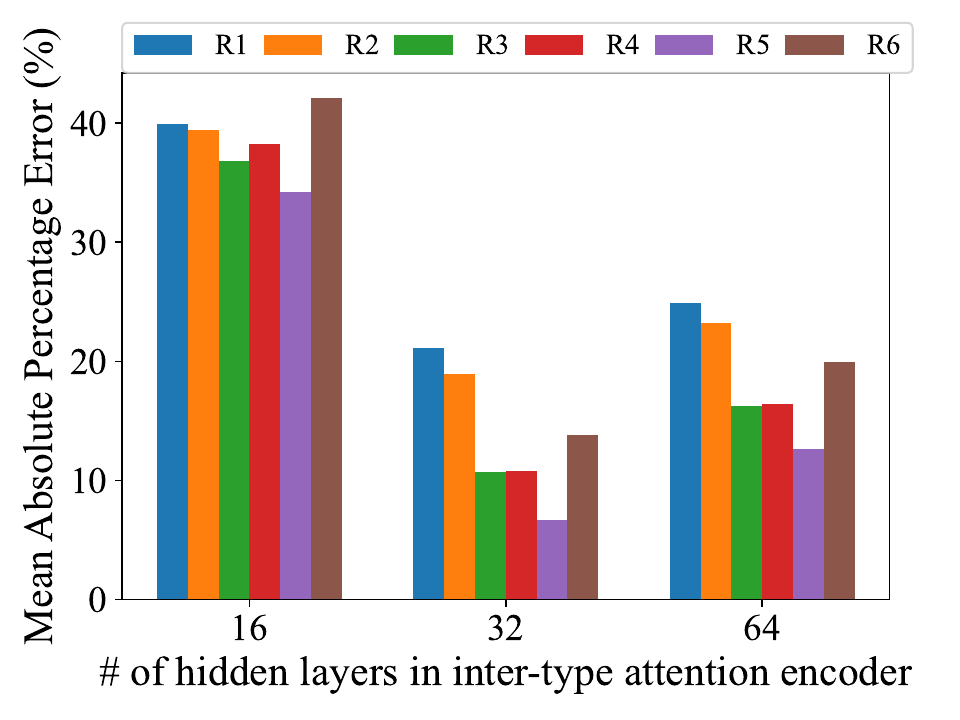}
        \vspace{-0.2in}
        \caption{Attention layer count impact}
        \label{fig:numberofattentionlayers}
    \end{minipage}
\hfill
        \begin{minipage}[b]{0.32\linewidth}
        \centering
            \includegraphics[width=2.3in]{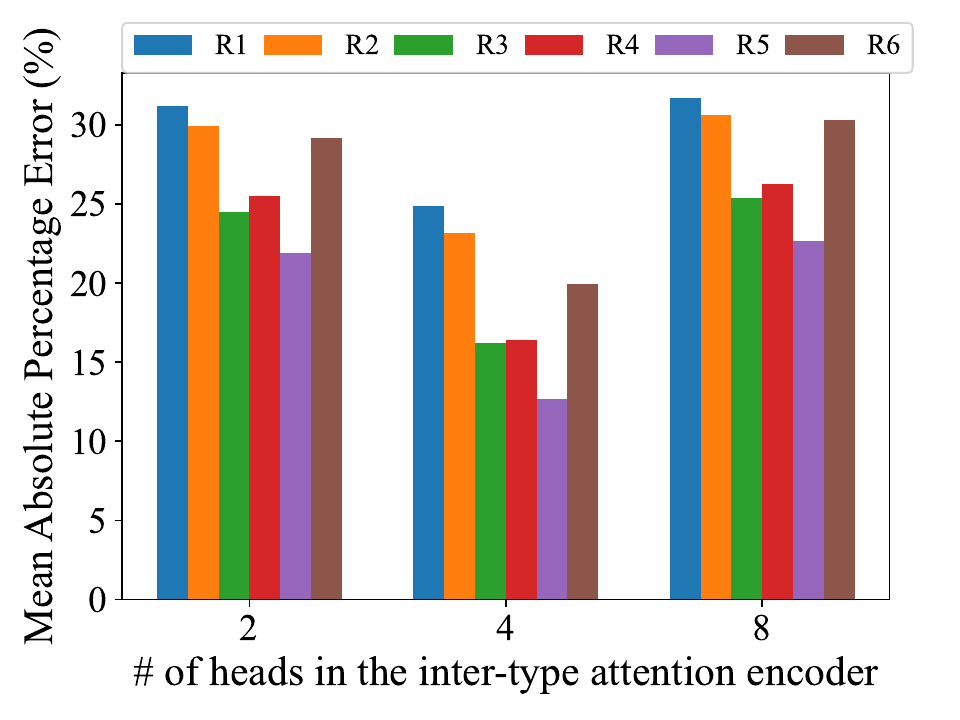}
            \vspace{-0.2in}
            \caption{Attention head count impact}
        \label{fig:numberofattentionheaders}
    \end{minipage}
      \hfill
    \begin{minipage}[b]{0.32\linewidth}
        \centering
        \includegraphics[width=2.3in]{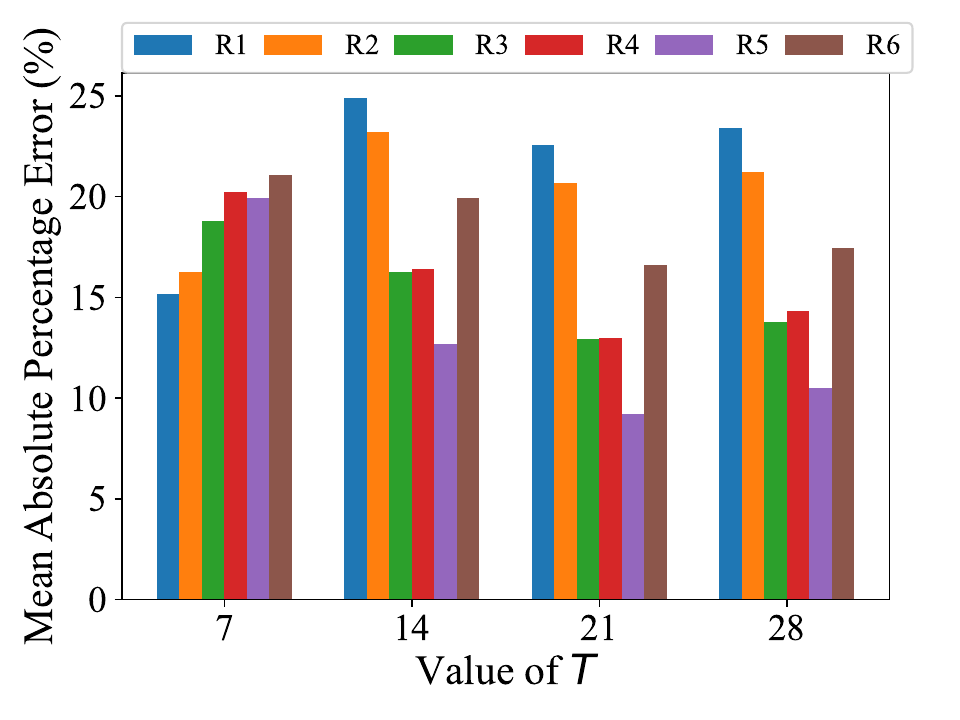}
        \vspace{-0.2in}
        \caption{Look-back window size impact}
        \label{fig:lookbackdays}
    \end{minipage}
    \vspace{-0.1in}
\end{figure*}

\begin{figure*}[h]
    \centering
        \begin{minipage}[b]{0.32\linewidth}
        \centering
        \includegraphics[width=2.3in]{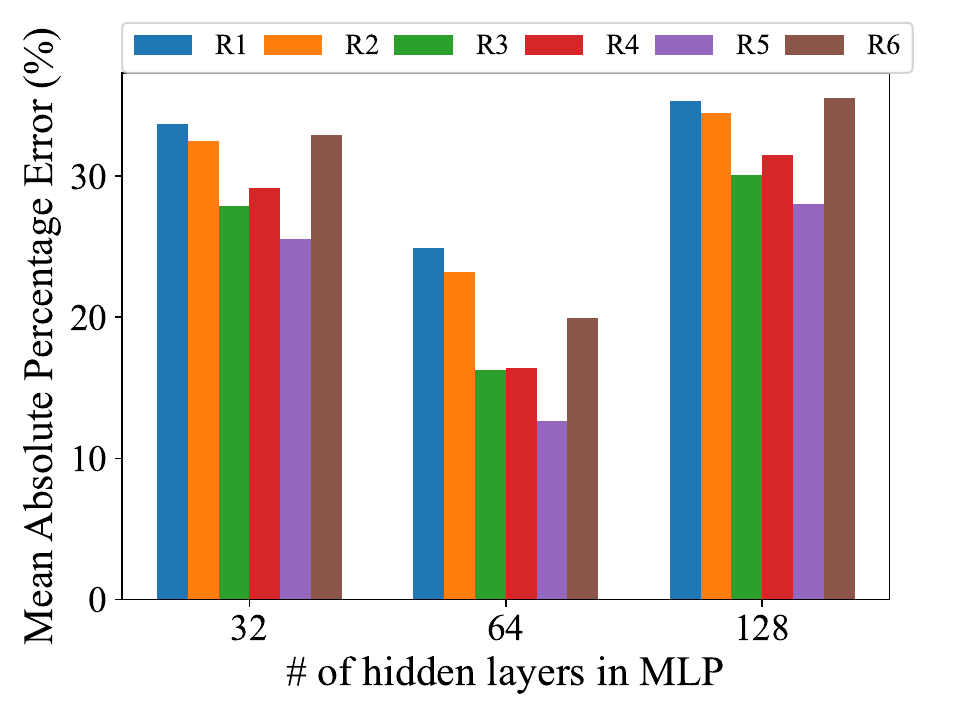}
        \vspace{-0.2in}
        \caption{MLP layer count impact}
        \label{fig:numberofmlplayers}
    \end{minipage}
    \hfill
        \begin{minipage}[b]{0.32\linewidth}
        \centering
            \includegraphics[width=2.3in]{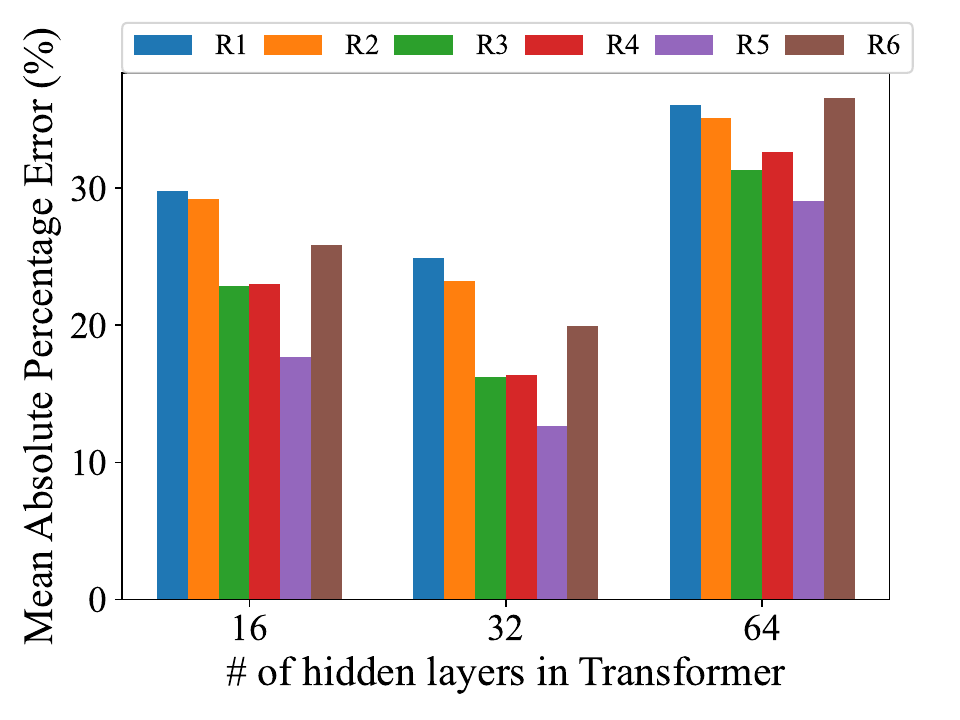}
            \vspace{-0.2in}
            \caption{Transformer layer count impact}
        \label{fig:numberoftransformerlayer}
    \end{minipage}
    \hfill
    \begin{minipage}[b]{0.32\linewidth}
        \centering
        \includegraphics[width=2.3in]{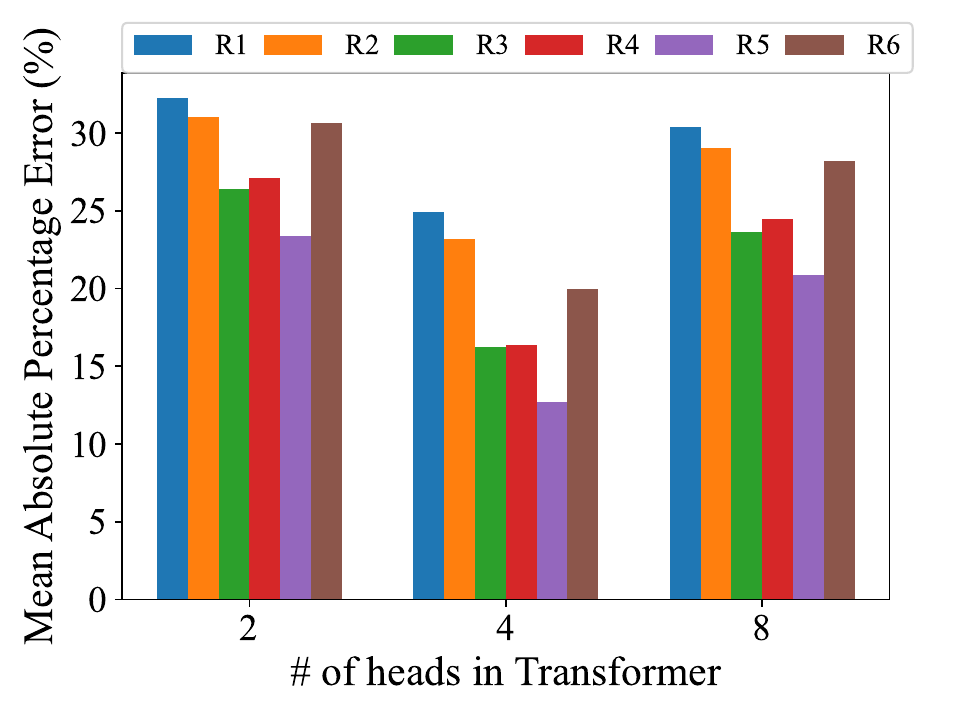}
        \vspace{-0.2in}
        \caption{Transformer head count impact}
        \label{fig:numberoftransformerheaders}
    \end{minipage}
    \vspace{-0.2in}
\end{figure*}

\vspace{-0.1in}
\subsection{Method Generalizability}

Table~\ref{table:newarkresults} presents the MAPE results for six request types across different methods using the Newark dataset. MuST$^2$-Learn consistently outperforms the baselines on most request types, with at least a 38.2\% reduction in MAPE. The only exception is R2 (i.e., Animal Complaint), where Prophet achieves slightly better performance.
A possible reason for this is that animal complaints often require specialized personnel and equipment that are not shared with other service types, reducing the benefit from inter-type modeling. Additionally, these requests may exhibit strong seasonal or holiday-related patterns, which Prophet captures more effectively due to its built-in seasonal modeling.
We also observe that deep neural network-based methods, such as Transformer, CNN, and IGTrans, perform poorly on this dataset. A likely reason is the limited number of samples, which makes it difficult to effectively train these models. In contrast, our method incorporates statistical distribution information through the intra-type variation module, which contributes to its improved performance under data-scarce conditions.
In summary, the results in Table~\ref{table:newarkresults} demonstrate that MuST$^2$-Learn generalizes well to a different dataset and maintains competitive accuracy.

\vspace{-0.1in}
\subsection{Ablation Study}

To further evaluate the effectiveness of the proposed method, we conduct an ablation study. We design the experiments to demonstrate the performance improvements resulting from integrating each component into the model.
Several variations of our method are introduced as:
\begin{itemize}[leftmargin=*]
    \item \textit{MuST$^2$-Learn-t}: This variant captures spatial, inter-type, and intra-type correlation without considering the temporal correlation.
    \item \textit{MuST$^2$-Learn-ct}: This variant captures temporal and intra-type correlation without considering the spatial and inter-type correlation.
    \item \textit{MuST$^2$-Learn-v}: This variant captures spatial, temporal and inter-type correlation without considering the intra-type variation correlation.
\end{itemize}

The ablation study results are presented in Fig.~\ref{fig:ablationstudy}. The full version of MuST$^2$-Learn consistently achieves the best performance across all request types in terms of MAPE. This highlights the importance of jointly modeling spatial, temporal, inter-type, and intra-type correlations for accurate service time prediction.

Among the components, the intra-type variation module has the most significant impact on performance. When this module is removed (MuST$^2$-Learn-v), the MAPE increases substantially, for example, from 24.90\% to 94.72\%. The temporal module, and the combined spatial and inter-type module, provide the second and third most important contributions, respectively, to the overall performance.

\vspace{-0.1in}
\subsection{Parameter Sensitivity Analysis}

We also analyze the impact of hyperparameters on the performance of MuST$^2$-Learn, and show the results measured by MAPE from  Fig.~\ref{fig:numberofattentionlayers} to Fig.~\ref{fig:numberoftransformerlayer}.
The descriptions of these figures are listed as follows.

\subsubsection{Inter-Type Attention Layers and Heads}
The experiments test 16, 32, and 64 hidden layers for the inter-type attention encoder and the results are shown in Fig.~\ref{fig:numberofattentionlayers}. The best performance in terms of MAPE is observed with 32 layers. This indicates that a moderate capacity in the attention layer is sufficient to capture the interaction across heterogeneous request types without overfitting. Similarly, as illustrated in Fig.~\ref{fig:numberofattentionheaders}, the number of attention heads (tested as 2, 4, and 8) showed that 4 heads performed best. Increasing to 8 heads likely led to redundancy and over-parameterization, while 2 heads may not have captured enough inter-type interaction.

\subsubsection{Temporal Window Length $T$}
The model's performance is also evaluated under varying values of 
$T$ (the length of the historical time window), with values of 7, 14, 21, and 28 days. As shown in Fig.~\ref{fig:lookbackdays}, the 21-day window results in the lowest prediction error, suggesting that service patterns require about three weeks of history to be effectively modeled. Shorter windows likely miss recurring patterns, while longer windows may have introduced noise or outdated context.

\subsubsection{Number of MLP Layers}
The number of hidden layers in the MLP is tested with values 32, 64, and 128. We demonstrate results in Fig.~\ref{fig:numberofmlplayers}. A setting of 64 hidden layers gives the best results, indicating a good trade-off between model complexity and training stability. The smaller 32-layer setting might lack sufficient capacity, and 128 layers might overfit the data, especially if the dataset size is limited.

\subsubsection{Transformer Encoder Layers and Heads}
For the Transformer encoder, both the hidden layer size and number of attention heads are explored. The results are shown in Fig.~\ref{fig:numberoftransformerlayer} and Fig.~\ref{fig:numberoftransformerheaders}. Similar to the inter-type attention module, 32 hidden layers and 4 heads provide the most accurate results. Larger settings increases MAPE, likely due to overfitting or vanishing gradients in deeper layers.

\section{Related Work}

We categorize the related work into three groups: municipal services, temporal prediction in urban applications, and multi-view learning.

\noindent \textbf{(i) Municipal Services}. 
With the deployment of 311 systems in major cities across Canada and the United States, many studies have focused on resident reporting patterns and service equity across regions~\cite{kontokosta2017equity,wu2020determinants,li2020311,kontokosta2021bias,xu2020closing,hsu2022towards,10.1145/3450267.3450538}, as well as service time forecasting~\cite{wang2022community,10.1145/3286978.3287010,raj2021swift}.
\cite{kontokosta2021bias} evaluates bias in resident-reported data by analyzing socio-spatial disparities in 311 complaint behavior, which requires more equitable service delivery.
\cite{xu2020closing} explores how 311 systems affect distributional equity in public service delivery with a finding that minority groups tend to utilize 311 systems to submit requests for essential service after disasters.
\cite{wang2022community} applies the Prophet model to estimate the response time of specific 311 service calls by incorporating trend, seasonality, and holiday effects.
\cite{10.1145/3286978.3287010} and \cite{raj2021swift} develop structured regression models based on sparse Gaussian Conditional Random Fields to predict response times.
This work advances prior studies on service time forecasting by integrating inter-type interactions among request types, capturing intra-type variation, and incorporating type-specific features. These enhancements contribute to improved prediction accuracy.

\vspace{3pt}
\noindent \textbf{(ii) Temporal Prediction in Urban Applications}. 
Temporal prediction plays a key role in urban applications, improving user experience in tasks such as delivery time and travel time estimation~\cite{zhao2023hst,zhou2023inductive,yi2023deepsta,zhang2024package,li2018multi,zhong2024adaptive}.
\cite{zhong2024adaptive} designs an adaptive cross-platform transportation time prediction framework for e-commerce orders using a hypergraph-based structure.
\cite{li2018multi} proposes a multi-task representation learning model to estimate arrival times based on origin, destination, and departure time.
\cite{zhao2023hst} introduces a heterogeneous spatial-temporal graph transformer for end-to-end delivery time prediction in warehouse-distribution systems.
To estimate package delivery time, \cite{zhou2023inductive} develops an inductive graph transformer that combines raw features with structural graph information.
These existing works primarily address temporal prediction tasks within a homogeneous context, such as package delivery or travel time estimation.
In contrast, this work considers a more complex setting involving heterogeneous types of service requests that interact with each other.
This added dimension of request heterogeneity motivates the new modeling approach proposed in this study.

\vspace{3pt}
\noindent \textbf{(iii) Multi-View Learning}.
Recently, there has been increasing attention on multi-view learning, which aims to learning representations from multiple views or data sources in different domains~\cite{10.1145/3308558.3313730,fang2024beyond,chang2025multi,lin2024mulste,10.1145/3637528.3672043,zhang2024multi}.
\cite{10.1145/3308558.3313730} proposes a multi-view and multi-modal spatial-temporal learning framework to forecast various categories of citywide abnormal events. In contrast, this work differs in two key aspects: (i) it addresses a regression task, whereas \cite{10.1145/3308558.3313730} focuses on a classification problem; 
(ii) the unique features of individual requests and the intra-type variation are not captured by \cite{10.1145/3308558.3313730}, as it does not account for differences among samples within the same type.
\cite{fang2024beyond} develops an ambiguity-aware multi-view learning framework to improve generalizability and reliability for unseen data beyond the training distribution.
In \cite{chang2025multi}, a multi-view knowledge representation learning framework is proposed for personalized news recommendation, integrating a multi-view news encoder with a candidate-aware attention mechanism.
However, both \cite{fang2024beyond} and \cite{chang2025multi} overlook the interactions between different views, such as the mutual influence among heterogeneous request types, which are critical for accurate service time estimation.
To jointly estimate purchase demand and delivery supply in e-commerce, \cite{lin2024mulste} introduces a multi-view spatio-temporal learning framework with heterogeneous event fusion.
Whereas our work additionally accounts for intra-view variation (i.e., intra-type variation among municipal service requests) and incorporates request-specific contextual information (i.e., text-based descriptions) in the learning process, which are significantly different from existing studies.

\vspace{-0.1in}
\section{Conclusion}
In this paper, we study the problem of estimating the service time of municipal services by tackling key challenges, including dynamic spatial-temporal correlations, underlying inter-type interactions, and high intra-type variations. We propose MuST$^2$-Learn, a multi-view spatial-temporal-type learning framework that incorporates dedicated encoders to capture these latent patterns across spatial, temporal, and type dimensions. We evaluate the proposed method on two real-world datasets collected from the City of Chattanooga and the City of Newark. Experimental results show that MuST$^2$-Learn outperforms state-of-the-art methods, reducing the mean absolute error by at least 32.5\%.

\section{Acknowledgement}
This publication was partially supported by the FY2024 and FY2025 Center of Excellence for Applied Computational Science competition in the University of Tennessee at Chattanooga.



\bibliographystyle{IEEEtran}
\bibliography{reference}

\begin{thebibliography}{10}
\providecommand{\url}[1]{#1}
\csname url@samestyle\endcsname
\providecommand{\newblock}{\relax}
\providecommand{\bibinfo}[2]{#2}
\providecommand{\BIBentrySTDinterwordspacing}{\spaceskip=0pt\relax}
\providecommand{\BIBentryALTinterwordstretchfactor}{4}
\providecommand{\BIBentryALTinterwordspacing}{\spaceskip=\fontdimen2\font plus
\BIBentryALTinterwordstretchfactor\fontdimen3\font minus \fontdimen4\font\relax}
\providecommand{\BIBforeignlanguage}[2]{{%
\expandafter\ifx\csname l@#1\endcsname\relax
\typeout{** WARNING: IEEEtran.bst: No hyphenation pattern has been}%
\typeout{** loaded for the language `#1'. Using the pattern for}%
\typeout{** the default language instead.}%
\else
\language=\csname l@#1\endcsname
\fi
#2}}
\providecommand{\BIBdecl}{\relax}
\BIBdecl

\bibitem{wang2017structure}
L.~Wang, C.~Qian, P.~Kats, C.~Kontokosta, and S.~Sobolevsky, ``Structure of 311 service requests as a signature of urban location,'' \emph{PloS one}, vol.~12, no.~10, p. e0186314, 2017.

\bibitem{nyc311}
\BIBentryALTinterwordspacing
NYC. Nyc311. [Online]. Available: \url{https://portal.311.nyc.gov/}
\BIBentrySTDinterwordspacing

\bibitem{chi311}
\BIBentryALTinterwordspacing
C.~of~Chicago. Chi311. [Online]. Available: \url{https://311.chicago.gov/s/?language=en_US}
\BIBentrySTDinterwordspacing

\bibitem{nam2014understanding}
T.~Nam and T.~A. Pardo, ``Understanding municipal service integration: An exploratory study of 311 contact centers,'' \emph{Journal of urban technology}, vol.~21, no.~1, pp. 57--78, 2014.

\bibitem{hashemi2022automatic}
M.~Hashemi, ``Automatic type detection of 311 service requests based on customer provided descriptions,'' \emph{Applied Artificial Intelligence}, vol.~36, no.~1, p. 2073717, 2022.

\bibitem{holmes2007building}
S.~T. Holmes, ``Building a 311 system: A case study of the orange county, florida, government service center,'' \emph{US Department of Justice, Office of Community Oriented Policing Services}, 2007.

\bibitem{wang2022community}
Y.~Wang, S.~R. Nagireddy, C.~T. Thota, D.~H. Ho, and Y.~Lee, ``Community-in-the-loop: Creating artificial process intelligence for co-production of city service,'' \emph{Proceedings of the ACM on Human-Computer Interaction}, vol.~6, no. CSCW2, pp. 1--21, 2022.

\bibitem{10.1145/3463677.3463717}
\BIBentryALTinterwordspacing
F.~Yusuf, S.~Cheng, S.~Ganapati, and G.~Narasimhan, ``Causal inference methods and their challenges: The case of 311 data,'' in \emph{Proceedings of the 22nd Annual International Conference on Digital Government Research}, ser. dg.o '21.\hskip 1em plus 0.5em minus 0.4em\relax New York, NY, USA: Association for Computing Machinery, 2021, p. 49–59. [Online]. Available: \url{https://doi.org/10.1145/3463677.3463717}
\BIBentrySTDinterwordspacing

\bibitem{zhao2023hst}
X.~Zhao, S.~Wang, H.~Wang, T.~He, D.~Zhang, and G.~Wang, ``Hst-gt: Heterogeneous spatial-temporal graph transformer for delivery time estimation in warehouse-distribution integration e-commerce,'' in \emph{Proceedings of the 32nd ACM International Conference on Information and Knowledge Management}, 2023, pp. 3402--3411.

\bibitem{10.1145/3286978.3287010}
\BIBentryALTinterwordspacing
D.~DeFazio, A.~Ramesh, and A.~Seetharam, ``Nycer: A non-emergency response predictor for nyc using sparse gaussian conditional random fields,'' in \emph{Proceedings of the 15th EAI International Conference on Mobile and Ubiquitous Systems: Computing, Networking and Services}, ser. MobiQuitous '18.\hskip 1em plus 0.5em minus 0.4em\relax New York, NY, USA: Association for Computing Machinery, 2018, p. 187–196. [Online]. Available: \url{https://doi.org/10.1145/3286978.3287010}
\BIBentrySTDinterwordspacing

\bibitem{raj2021swift}
R.~Raj, A.~Ramesh, A.~Seetharam, and D.~DeFazio, ``Swift: A non-emergency response prediction system using sparse gaussian conditional random fields,'' \emph{Pervasive and Mobile Computing}, vol.~71, p. 101317, 2021.

\bibitem{9291583}
G.~Bejarano, A.~Kulkarni, X.~Luo, A.~Seetharam, and A.~Ramesh, ``Deeper: A deep learning based emergency resolution time prediction system,'' in \emph{2020 International Conferences on Internet of Things (iThings) and IEEE Green Computing and Communications (GreenCom) and IEEE Cyber, Physical and Social Computing (CPSCom) and IEEE Smart Data (SmartData) and IEEE Congress on Cybermatics (Cybermatics)}, 2020, pp. 490--497.

\bibitem{zhou2023inductive}
X.~Zhou, J.~Wang, Y.~Liu, X.~Wu, Z.~Shen, and C.~Leung, ``Inductive graph transformer for delivery time estimation,'' in \emph{Proceedings of the Sixteenth ACM International Conference on Web Search and Data Mining}, 2023, pp. 679--687.

\bibitem{yi2023deepsta}
J.~Yi, H.~Yan, H.~Wang, J.~Yuan, and Y.~Li, ``Deepsta: A spatial-temporal attention network for logistics delivery timely rate prediction in anomaly conditions,'' in \emph{Proceedings of the 32nd ACM International Conference on Information and Knowledge Management}, 2023, pp. 4916--4922.

\bibitem{cohen2009pearson}
I.~Cohen, Y.~Huang, J.~Chen, J.~Benesty, J.~Benesty, J.~Chen, Y.~Huang, and I.~Cohen, ``Pearson correlation coefficient,'' \emph{Noise reduction in speech processing}, pp. 1--4, 2009.

\bibitem{vaswani2017attention}
A.~Vaswani, N.~Shazeer, N.~Parmar, J.~Uszkoreit, L.~Jones, A.~N. Gomez, {\L}.~Kaiser, and I.~Polosukhin, ``Attention is all you need,'' \emph{Advances in neural information processing systems}, vol.~30, 2017.

\bibitem{shah2025interference}
S.~L. Shah, N.~H. Mahmood, and M.~Latva-aho, ``Interference prediction using gaussian process regression and management framework for critical services in local 6g networks,'' in \emph{IEEE Wireless Communications and Networking Conference (WCNC)}.\hskip 1em plus 0.5em minus 0.4em\relax Mico Milano Congressi, Milan, Italy: IEEE, March 2025.

\bibitem{8317796}
H.~Rodriguez-Deniz, E.~Jenelius, and M.~Villani, ``Urban network travel time prediction via online multi-output gaussian process regression,'' in \emph{2017 IEEE 20th International Conference on Intelligent Transportation Systems (ITSC)}, 2017, pp. 1--6.

\bibitem{Roberts2013Gaussian}
\BIBentryALTinterwordspacing
S.~Roberts, M.~Osborne, M.~Ebden, S.~Reece, N.~Gibson, and S.~Aigrain, ``Gaussian processes for time-series modelling,'' \emph{Philosophical Transactions of The Royal Society A: Mathematical, Physical and Engineering Sciences}, vol. 371, no. 1984, p. 20110550, 2013. [Online]. Available: \url{https://doi.org/10.1098/rsta.2011.0550}
\BIBentrySTDinterwordspacing

\bibitem{Arora2023review}
\BIBentryALTinterwordspacing
G.~Arora, K.~Bala, H.~Emadifar, and M.~Khademi, ``A review of radial basis function with applications explored,'' \emph{Journal of the Egyptian Mathematical Society}, vol.~31, no.~1, p.~6, 2023. [Online]. Available: \url{https://joems.springeropen.com/articles/10.1186/s42787-023-00164-3}
\BIBentrySTDinterwordspacing

\bibitem{matallm}
\BIBentryALTinterwordspacing
Meta. Llama 4: Leading intelligence. unrivaled speed and efficiency. [Online]. Available: \url{https://www.llama.com/}
\BIBentrySTDinterwordspacing

\bibitem{newark}
\BIBentryALTinterwordspacing
City of newark 311. [Online]. Available: \url{https://seeclickfix.com/newark}
\BIBentrySTDinterwordspacing

\bibitem{shumway2017arima}
R.~H. Shumway, D.~S. Stoffer, R.~H. Shumway, and D.~S. Stoffer, ``Arima models,'' \emph{Time series analysis and its applications: with R examples}, pp. 75--163, 2017.

\bibitem{ige2024state}
A.~O. Ige and M.~Sibiya, ``State-of-the-art in 1d convolutional neural networks: A survey,'' \emph{IEEE Access}, 2024.

\bibitem{kontokosta2017equity}
C.~Kontokosta, B.~Hong, and K.~Korsberg, ``Equity in 311 reporting: Understanding socio-spatial differentials in the propensity to complain,'' \emph{arXiv preprint arXiv:1710.02452}, 2017.

\bibitem{wu2020determinants}
W.-N. Wu, ``Determinants of citizen-generated data in a smart city: Analysis of 311 system user behavior,'' \emph{Sustainable Cities and Society}, vol.~59, p. 102167, 2020.

\bibitem{li2020311}
Y.~Li, A.~Hyder, L.~T. Southerland, G.~Hammond, A.~Porr, and H.~J. Miller, ``311 service requests as indicators of neighborhood distress and opioid use disorder,'' \emph{Scientific reports}, vol.~10, no.~1, p. 19579, 2020.

\bibitem{kontokosta2021bias}
C.~E. Kontokosta and B.~Hong, ``Bias in smart city governance: How socio-spatial disparities in 311 complaint behavior impact the fairness of data-driven decisions,'' \emph{Sustainable Cities and Society}, vol.~64, p. 102503, 2021.

\bibitem{xu2020closing}
C.~K. Xu and T.~Tang, ``Closing the gap or widening the divide: The impacts of technology-enabled coproduction on equity in public service delivery,'' \emph{Public administration review}, vol.~80, no.~6, pp. 962--975, 2020.

\bibitem{hsu2022towards}
J.~H.-P. Hsu, J.~Wang, and M.~Lee, ``Towards an expectation-oriented model of public service quality: A preliminary study of nyc 311,'' in \emph{International Conference on Social Informatics}.\hskip 1em plus 0.5em minus 0.4em\relax Springer, 2022, pp. 447--458.

\bibitem{10.1145/3450267.3450538}
\BIBentryALTinterwordspacing
Y.~Yuan, M.~Ma, S.~Han, D.~Zhang, F.~Miao, J.~Stankovic, and S.~Lin, ``Deresolver: a decentralized negotiation and conflict resolution framework for smart city services,'' in \emph{Proceedings of the ACM/IEEE 12th International Conference on Cyber-Physical Systems}, ser. ICCPS '21.\hskip 1em plus 0.5em minus 0.4em\relax ACM, 2021, p. 98–109. [Online]. Available: \url{https://doi.org/10.1145/3450267.3450538}
\BIBentrySTDinterwordspacing

\bibitem{zhang2024package}
L.~Zhang, Y.~Liu, Z.~Zeng, Y.~Cao, X.~Wu, Y.~Xu, Z.~Shen, and L.~Cui, ``Package arrival time prediction via knowledge distillation graph neural network,'' \emph{ACM Transactions on Knowledge Discovery from Data}, vol.~18, no.~5, pp. 1--19, 2024.

\bibitem{li2018multi}
Y.~Li, K.~Fu, Z.~Wang, C.~Shahabi, J.~Ye, and Y.~Liu, ``Multi-task representation learning for travel time estimation,'' in \emph{Proceedings of the 24th ACM SIGKDD international conference on knowledge discovery \& data mining}, 2018, pp. 1695--1704.

\bibitem{zhong2024adaptive}
S.~Zhong, W.~Lyu, Z.~Hong, G.~Yang, W.~Zuo, H.~Wang, G.~Wang, Y.~Yang, and D.~Zhang, ``Adaptive cross-platform transportation time prediction for logistics,'' in \emph{Proceedings of the 33rd ACM International Conference on Information and Knowledge Management}, 2024, pp. 5127--5134.

\bibitem{10.1145/3308558.3313730}
\BIBentryALTinterwordspacing
C.~Huang, C.~Zhang, J.~Zhao, X.~Wu, D.~Yin, and N.~Chawla, ``Mist: A multiview and multimodal spatial-temporal learning framework for citywide abnormal event forecasting,'' in \emph{The World Wide Web Conference}, ser. WWW '19.\hskip 1em plus 0.5em minus 0.4em\relax New York, NY, USA: Association for Computing Machinery, 2019, p. 717–728. [Online]. Available: \url{https://doi.org/10.1145/3308558.3313730}
\BIBentrySTDinterwordspacing

\bibitem{fang2024beyond}
Z.~Fang, S.~Du, Y.~Chen, and S.~Wang, ``Beyond the known: Ambiguity-aware multi-view learning,'' in \emph{Proceedings of the 32nd ACM International Conference on Multimedia}, 2024, pp. 8518--8526.

\bibitem{chang2025multi}
C.~Chang, F.~Tang, P.~Yang, J.~Zhang, J.~Huang, J.~Li, and Z.~Li, ``Multi-view knowledge representation learning for personalized news recommendation,'' \emph{Scientific Reports}, vol.~15, no.~1, p. 1152, 2025.

\bibitem{lin2024mulste}
L.~Lin, Z.~Lu, S.~Wang, Y.~Liu, Z.~Hong, H.~Wang, and S.~Wang, ``Mulste: A multi-view spatio-temporal learning framework with heterogeneous event fusion for demand-supply prediction,'' in \emph{Proceedings of the 30th ACM SIGKDD Conference on Knowledge Discovery and Data Mining}, 2024, pp. 1781--1792.

\bibitem{10.1145/3637528.3672043}
\BIBentryALTinterwordspacing
Y.~Luo, K.~Yang, M.~Hong, X.~Y. Liu, Z.~Nie, H.~Zhou, and Z.~Nie, ``Learning multi-view molecular representations with structured and unstructured knowledge,'' in \emph{Proceedings of the 30th ACM SIGKDD Conference on Knowledge Discovery and Data Mining}, ser. KDD '24.\hskip 1em plus 0.5em minus 0.4em\relax New York, NY, USA: Association for Computing Machinery, 2024, p. 2082–2093. [Online]. Available: \url{https://doi.org/10.1145/3637528.3672043}
\BIBentrySTDinterwordspacing

\bibitem{zhang2024multi}
Y.~Zhang, Y.~Feng, W.-J. Zhou, Y.~Ye, M.~Tan, R.~Xiao, H.~Tang, J.~Ding, and J.~Yu, ``Multi-domain deep learning from a multi-view perspective for cross-border e-commerce search,'' in \emph{Proceedings of the AAAI Conference on Artificial Intelligence}, vol.~38, no.~8, 2024, pp. 9387--9395.

\end{thebibliography}

\end{document}